\documentclass[journal,twoside,web]{ieeecolor}
\usepackage{generic}
\usepackage{cite}

\usepackage{amsmath,amssymb,amsfonts}
\usepackage{array}
\usepackage{booktabs}
\usepackage{graphicx}
\usepackage{algorithm,algorithmic}
\usepackage{hyperref}
\hypersetup{hidelinks=true}

\usepackage{multirow}
\usepackage{pifont}

\usepackage{textcomp}
\def\BibTeX{{\rm B\kern-.05em{\sc i\kern-.025em b}\kern-.08em
    T\kern-.1667em\lower.7ex\hbox{E}\kern-.125emX}}
\begin{document}
\title{Label-Efficient Deep Learning for ECG Delineation: A Multi-Dataset Benchmark against Widely Used Delineation Tools}
\author{Jeonghwa Lim$^*$, Minje Park$^*$, Yeongyeon Na, Yujin Eom, Soyeon Lim, Young Ho Lee, Yu Jeong Kim, Sunghoon Joo$^\dagger$, and Ki Hong Lee$^\ddagger$
\thanks{This research was supported by a grant of the Korea Health Technology R\&D Project through the Korea Health Industry Development Institute (KHIDI), funded by the Ministry of Health \& Welfare, Republic of Korea (grant number: RS-2023-00265466).}
\thanks{$^*$ Both authors contributed equally to this research.}
\thanks{J. Lim, M. Park, and  Y. Na are with VUNO Inc., Seoul, South Korea (e-mail: \{jeonghwa.lim, minje.park, yeongyeon\}@vuno.co).}
\thanks{Y. Eom is with the Department of AI Mobility Engineering, Ajou University, Suwon, South Korea (e-mail: eomyj2@ajou.ac.kr).} 
\thanks{S. Lim is with C\&Thoth Co., Ltd., Seoul, South Korea (e-mail: sylim@cnthoth.com).}
\thanks{Y. H. Lee is with the Department of Biomedical Sciences, Chonnam National University Graduate School, and the Department of Cardiovascular Medicine, Chonnam National University Hospital, Gwangju, South Korea (e-mail: onuri205dev@gmail.com).}
\thanks{Y, J, Kim is with the Department of Cardiovascular Medicine, Chonnam National University Hospital, Gwangju, South Korea (e-mail: rmfls925@gmail.com).}
\thanks{$^\dagger$ Corresponding author. S. Joo is with VUNO Inc., Seoul, South Korea (e-mail: sunghoon.joo@vuno.co).}
\thanks{$^\ddagger$ Corresponding author. K. H. Lee is with the Department of Cardiovascular Medicine, Chonnam National University Hospital, and the Department of Internal Medicine, Chonnam National University Medical School, Gwangju, South Korea (e-mail: drgood2@naver.com).}
}

\maketitle

\begin{abstract}
    Electrocardiogram (ECG) delineation, the identification of waveform boundaries, is a foundational step that translates raw ECG signals into clinically interpretable measurements. Deep learning has advanced this task but remains dependent on costly expert annotations. Label-efficient strategies such as self-supervised pretraining and semi-supervised learning are expected to ease this burden, yet it remains unclear whether they yield reliable delineation and whether the deep models they produce outperform the delineation tools used in practice. We address this in two stages. First, comparing self-supervised objectives with supervised or semi-supervised fine-tuning across one internal and four external datasets, we find that pretraining helps but the objective matters, and that the value of semi-supervised fine-tuning depends on the pretraining objective. Second, we benchmark the selected deep learning model against widely used open-source (\textsf{NeuroKit2}, \textsf{Prominence}, \textsf{ECGdeli}) and commercial (\textsf{CalECG}) tools using three complementary metrics. The model ranks best on every metric and dataset, outperforming the strongest tool by a clear margin on the rhythm-diverse set (mIoU 71.3 vs. 54.8\%; averaged point-wise sensitivity 92.6 vs. 76.4\%), and degrades the least from sinus to arrhythmia. A rhythm-stratified and point-wise analysis further characterizes the distinctive behavior of each tool, yielding practical guidance for tool selection. These results provide systematic, multi-dataset evidence that self-supervised pretraining is effective for ECG delineation and enables a label-efficiently trained deep learning model to outperform widely used delineation tools by leveraging abundant unlabeled data. This supports adopting such models in diverse, real-world clinical settings.
\end{abstract}

\begin{IEEEkeywords}
Electrocardiography, Semantic segmentation, Electrocardiogram delineation, Label-efficient learning, Self-supervised learning, Semisupervised learning, Benchmarking
\end{IEEEkeywords}

\section{Introduction}
\label{sec:introduction}
\IEEEPARstart{A}{n} Electrocardiogram (ECG) delineation, the segmentation of the ECG signal into its component P wave, QRS complex, and T wave, forms the basis for deriving clinically essential measurements such as the PR interval, QRS duration, and QT interval~\cite{gacek2011ecg}. These intervals are fundamental to the diagnosis and monitoring of a wide range of cardiac conditions, including conduction abnormalities, arrhythmias, and other electrophysiologic disorders such as QT prolongation~\cite{rautaharju2009aha}. Accordingly, accurate and reliable ECG delineation is essential, as it directly affects the quality of downstream clinical decision-making~\cite{kligfield2007recommendations}. Yet achieving robust, generalizable ECG delineation remains challenging, owing to substantial morphological variability and signal-quality degradation across patients and recording conditions~\cite{elgendi2014revisiting, liu2018performance}.

Automated delineation has followed two broad approaches: rule-based signal processing and deep learning. Rule-based methods are interpretable, lightweight, and widely deployed, but their accuracy tends to degrade on morphologically atypical beats and under residual noise, such as baseline wander and muscle artifacts, that distorts the waveform features on which they rely~\cite{elgendi2014revisiting, liu2018performance}. Recently, deep learning has been increasingly applied to delineation, learning waveform morphology directly from data and achieving improved accuracy~\cite{jimenez2019u,liang2022ecg_segnet,joung2024deep}. The key limitation of deep learning, however, is a reliance on large volumes of expert-annotated data: precise delineation labels are costly and time-consuming to obtain, which constrains the practical scalability of fully supervised training.

\begin{figure*}[t]
    \centering
    \includegraphics[width=0.9\linewidth]{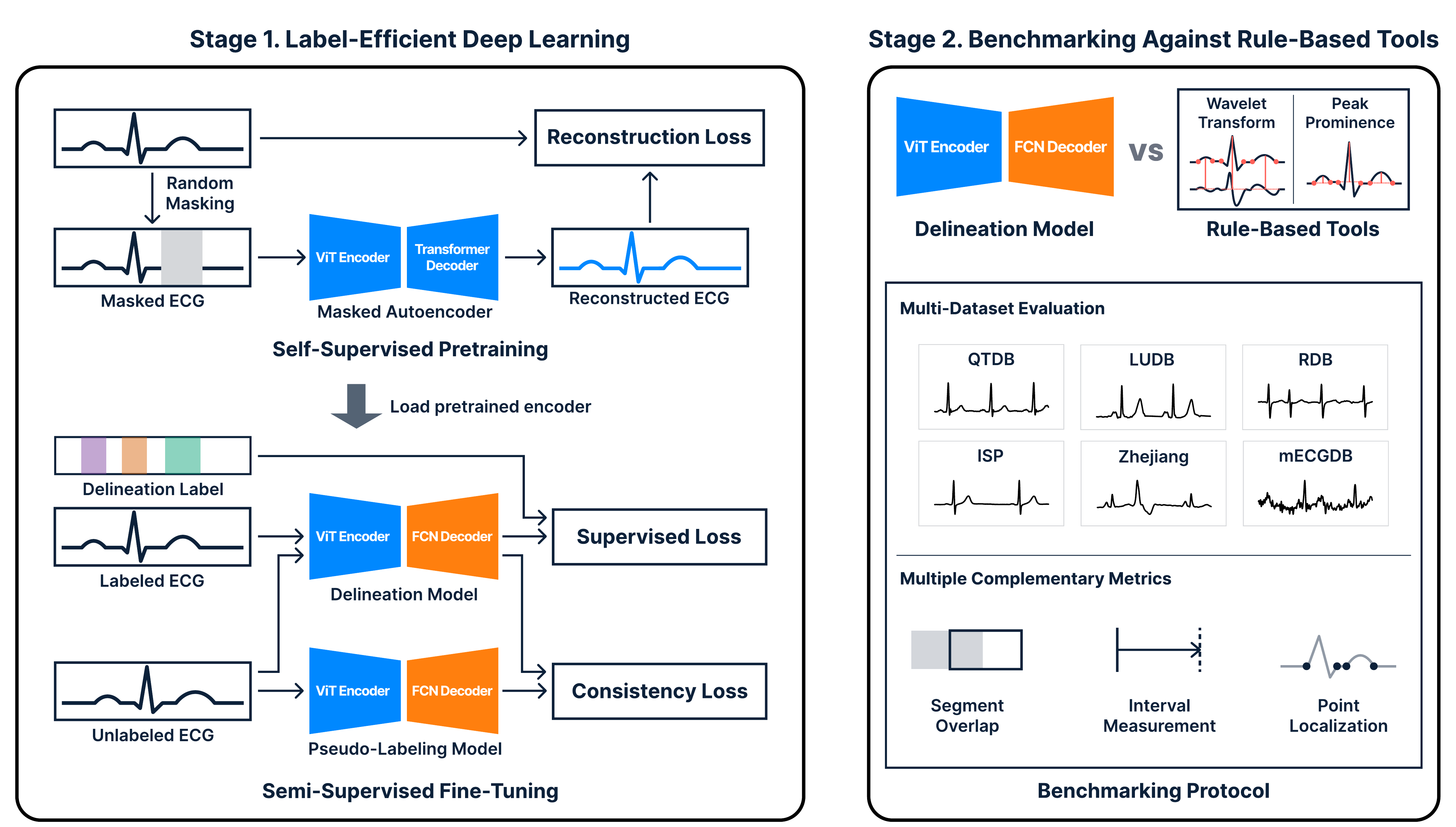}
    \caption{
        Study overview. Stage 1 pretrains an encoder on unlabeled ECG data with self-supervised learning (e.g., Masked Autoencoder) and fine-tunes it with a segmentation decoder (e.g., FCN) under supervised and semi-supervised learning (e.g., Mean Teacher). Stage 2 benchmarks the resulting model against existing delineation tools on multiple datasets using three complementary metrics: mIoU, measurement error on the PR, QRS, and QT intervals, and fiducial point localization accuracy.
    }
    \label{fig:main}
\end{figure*}

Label-efficient deep learning~\cite{jin2023label} can be a promising approach to mitigate the annotation bottleneck.
Semi-supervised learning combines a limited labeled set with abundant unlabeled data, improving over supervised training on the labeled subset alone~\cite{pelaez2023survey}.
Self-supervised pretraining, which learns transferable representations from large-scale unlabeled data before task-specific fine-tuning, has driven progress across various domains~\cite{krishnan2022self,goldberger2000physiobank}, including ECG tasks such as arrhythmia detection and disease screening~\cite{kiyasseh2021clocs,na2024guiding,liu2024zero}.
The effectiveness of label-efficient learning for ECG delineation, however, remains unexamined on two fronts.
First, ECG delineation is a dense prediction task at the level of individual time points, unlike the record-level classification tasks where pretraining has proven effective~\cite{al2026benchmarking}.
It is therefore unclear whether self-supervised pretraining helps at all, how different pretraining objectives compare, and how the choice of objective interacts with supervised versus semi-supervised fine-tuning.
Second, setting aside how these models are trained, label-efficient learning approaches have, to our knowledge, never been directly benchmarked against the open-source and commercial tools currently used in research and clinical practice.

We address these questions in two stages, illustrated in Fig.~\ref{fig:main}. In the first stage, we compare several pretraining objectives, each followed by supervised or semi-supervised fine-tuning, to assess the effect of self-supervised pretraining and the fine-tuning strategy on delineation performance. In the second stage, we benchmark the model selected in the first stage against established open-source and commercial tools.

Our contributions are summarized as follows:
\begin{itemize}
\item We provide systematic, multi-dataset evidence that self-supervised pretraining benefits ECG delineation.
\item We present the first benchmark of label-efficiently trained deep learning models against multiple open-source libraries as well as a commercial package, with rhythm-stratified analysis across diverse external datasets.
\item We characterize the distinctive behavior of each tool across datasets, and show that the deep learning model achieves the best overall performance while remaining robust under arrhythmia.
\end{itemize}

\section{Related Works}
\subsection{Classical ECG Delineation Tools}
Rule-based signal processing has been the standard approach to automated ECG delineation.
Among these methods, the dominant family builds on the wavelet transform, which decomposes the signal into multiple scales so that the peaks and boundaries of the P wave, QRS complex, and T wave can be localized from characteristic slopes~\cite{martinez2004wavelet}.
Many of these algorithms are distributed as open-source and are widely used in practice.
\textsf{NeuroKit2}~\cite{makowski2021neurokit2} and \textsf{ECGdeli}~\cite{pilia2021ecgdeli} implement wavelet-based delineation, while \textsf{Prominence}~\cite{Emrich2024} is a recent physiology-informed method that localizes fiducial points via peak prominence with linear runtime.
In regulatory and pharmaceutical settings, commercial packages such as \textsf{CalECG} (AMPS LLC, New York, NY) instead produce an initial delineation that a trained operator reviews and corrects~\cite{porretta2025software, engstrom2022validation}.

\begin{table*}[t]
\centering
\footnotesize
\setlength{\tabcolsep}{6pt}
\renewcommand{\arraystretch}{1.2}
\caption{Characteristics of the ECG datasets used for developing and evaluating the delineation model. Lead configurations range from 12-lead (six limb and six precordial leads) to 6-lead (limb only) and 2-lead (arbitrary pairs, e.g., MLII and V1).}
\label{tab:datasets}
\begin{tabular}{@{}l l rr l r l r@{}}
\toprule
Role & Source & \#Subjects & \#ECGs & Duration (labeled) & Sample rate & Lead type & \#Samples \\
\midrule
\multirow{3}{*}{Training}
 & \textit{QTDB} & 105 & 105 & 5.9--253.6 s & 250 Hz & 2-lead & 718 \\
 & \textit{ISP}  & 499 & 499 & 10 s & 1000 Hz & 12-lead & 5,988 \\
 & \textit{PTB-XL} & 18,885 & 21,837 & --- (unlabeled) & 500 Hz & 12-lead & 262,044 \\
\midrule
\multirow{4}{*}{Evaluation}
 & \textit{LUDB}     & 200   & 200   & 10 s & 500 Hz & 12-lead & 2,369 \\
 & \textit{Zhejiang} & 334   & 334   & 1.3--7.1 s & 2000 Hz & 12-lead & 4,008 \\
 & \textit{RDB}      & 2,399 & 2,399 & 10 s & 500 Hz & 12-lead & 28,788 \\
 & \textit{mECGDB}   & 205   & 205   & 2--10 s & 250 Hz & 6-lead & 205 \\
\bottomrule
\end{tabular}
\end{table*}

\subsection{Deep Learning for ECG Delineation}
Deep learning approaches formulate ECG delineation as a dense prediction task, assigning each time point to a wave class (P, QRS, T, or background) with encoder-decoder segmentation networks~\cite{jimenez2019u,liang2022ecg_segnet,joung2024deep}.
These models surpass rule-based methods in accuracy and extend to diverse arrhythmias~\cite{joung2024deep}, but they are predominantly supervised and depend on dense, time-point-level annotations that are costly to produce, especially across the varied morphologies seen in arrhythmia.
Semi-supervised learning offers a natural solution, supplementing a small labeled set with abundant unlabeled data~\cite{pelaez2023survey}.
A recent multi-dataset study applies it to delineation and reports gains over supervised training on the labeled subset alone~\cite{park2025semisegecg}, confirming that unlabeled recordings are a useful resource.

\subsection{Self-Supervised Learning}
Self-supervised learning (SSL) reduces the reliance on manual annotations by pretraining on large unlabeled datasets through pretext tasks constructed from the data itself, yielding general-purpose representations. These representations can then be transferred to the downstream task using only a small labeled set~\cite{chen2020simple,kiyasseh2021clocs}. Several categories of SSL methods exist, including reconstruction-based and contrastive approaches~\cite{he2022masked,he2020momentum,liu2024zero}. Although SSL has been widely applied to ECG, existing studies have largely targeted classification tasks such as arrhythmia detection and disease screening~\cite{kiyasseh2021clocs,na2024guiding,liu2024zero}; to our knowledge, none has treated delineation as a downstream task. Consequently, whether SSL benefits delineation and how pretraining interacts with supervised versus semi-supervised fine-tuning all remain underexamined.

\subsection{Benchmarking Gaps in ECG Delineation}
Jaeger et al. benchmarked rule-based methods across multiple datasets but excluded deep learning~\cite{jaeger2024systematic}, whereas Chuma et al. included deep learning methods but evaluated on only a single external dataset~\cite{chuma2026performance}.
Both assessed only waveform-level accuracy at the fiducial point or segment level, without reporting the clinical interval measurements (PR interval, QRS duration, and QT interval) that delineation ultimately serves.
A recent semi-supervised benchmark~\cite{park2025semisegecg} did report such interval metrics, but evaluated deep learning models alone and, by merging the labeled datasets into a single training pool with only one held-out external set, left generalization to unseen datasets underexplored.
To our knowledge, no prior work has jointly evaluated rule-based and deep learning methods across multiple external datasets while assessing these clinical intervals.

\section{Benchmark Design}
\label{sec:benchmarkd_design}
 
\subsection{Datasets and Preprocessing}
\label{sec:datasets}
We use seven public datasets, namely \textit{QTDB}~\cite{laguna1997database,jimenez2024generalising}, \textit{ISP}~\cite{avetisyan2024ispdataset}, \textit{PTB-XL}~\cite{wagner2020ptb}, \textit{MIMIC-IV-ECG}~\cite{gow2023mimic}, \textit{LUDB}~\cite{kalyakulina2020ludb}, \textit{Zhejiang}~\cite{zheng202012,jimenez2024generalising}, and \textit{RDB}~\cite{RDB202012,liu2025net}, together with a private dataset, the mobile ECG database (\textit{mECGDB}). To ensure a fair and reproducible comparison, we follow the standardized benchmark configuration of a prior open benchmark study~\cite{park2025semisegecg}. \textit{RDB} and \textit{MIMIC-IV-ECG}, which are not included in the benchmark, were processed following the same protocol. Table~\ref{tab:datasets} summarizes the characteristics of these datasets, and further details are provided in the Supplementary Material (Section~\ref{sec:supple_dataset_detail}).

\subsubsection{Pretraining data}
For self-supervised pretraining, we use \textit{MIMIC-IV-ECG}, a dataset that provides large-scale raw ECG signals alongside clinical text reports and has been widely adopted for self-supervised ECG representation learning~\cite{han2026review}. The original dataset contains 800,035 diagnostic 12-lead ECGs from 161,352 subjects. After filtering out corrupted recordings (e.g., those with saturated or NaN values) and those whose report is empty or contains fewer than three words~\cite{liu2024zero}, we retain 761,139 recordings from 154,094 subjects. Each recording is 10 seconds long and is sampled at 500 Hz.

\subsubsection{Training data}
Three datasets are used to develop the delineation model: \textit{QTDB} and \textit{ISP} as the labeled sets and \textit{PTB-XL} as the unlabeled set.
We choose \textit{QTDB} and \textit{ISP} for their complementary characteristics:
\textit{QTDB} is an established delineation benchmark, while \textit{ISP} offers larger scale, more realistic noise, and greater rhythm diversity.
They are split independently into training, validation, and test sets and merged into a single labeled pool (\textit{QTDB}$+$\textit{ISP}). Its training split provides the labeled data for supervised and semi-supervised training.
\textit{PTB-XL} is a widely used clinical 12-lead ECG dataset spanning diverse morphologies and rhythms. We therefore use it as the unlabeled set for semi-supervised training.

\subsubsection{Evaluation data}
The \textit{QTDB} and \textit{ISP} test splits together form our internal test set.
We additionally use four external test sets, chosen to span different sources, acquisition devices, and populations, so that robustness can be assessed under the diverse distribution shifts arising in practice.
Each external set contributes a distinct aspect of evaluation.
\textit{LUDB} is a widely used benchmark for ECG delineation, providing high-quality, per-lead annotations.
\textit{Zhejiang} consists of recordings that each contain arrhythmic beats, predominantly premature ventricular complexes (PVCs), enabling delineation assessment under arrhythmic conditions.
\textit{RDB} offers delineation labels across eight rhythm types, from sinus rhythms to various arrhythmias, supporting rhythm-stratified subgroup analysis.
Finally, \textit{mECGDB} comprises recordings collected with a portable device outside clinical settings, testing robustness under less controlled conditions.
All external sets provide dense time-point-level delineation labels, except \textit{mECGDB}, which offers only interval-level annotations (PR interval, QRS duration, and QT interval). The collection of \textit{mECGDB} was approved by the Institutional Review Board (IRB) of Chonnam National University Hospital (IRB number: CNUH-2025-127), and informed consent was obtained from all participants.

\subsubsection{Preprocessing}
Each ECG is adjusted to a fixed 10-second duration by cropping or zero-padding as needed, resampled to 250~Hz, band-pass filtered (0.67–40~Hz) to suppress baseline wander and high-frequency noise, and Z-score normalized before being passed to the models.
 
\subsection{Evaluation Protocol}
\label{sec:metrics}
We evaluate all methods with three complementary metrics: mean Intersection over Union (mIoU), an averaged interval error, and averaged point-wise sensitivity.
First, mIoU quantifies the overlap between predicted and reference segments across the P, QRS, and T classes, reflecting segmentation accuracy per time point. It is computed per sample and then averaged across the test set.
Second, for each clinical interval (PR interval, QRS duration, and QT interval), we compute its mean absolute error against the reference annotations, averaged over the samples where that interval is available; averaging the three per-interval errors yields a single averaged interval error, capturing the accuracy of the measurements used in clinical practice.
Third, we report point-wise sensitivity at the onset and offset of the P, QRS, and T waves, counting a predicted point as correct if it falls within a fixed tolerance of the reference. Sensitivity is conventionally reported per fiducial point, but for a concise comparison across methods and test sets, we report the average over the six onset and offset points and present the sensitivity for each point separately in the point-wise performance analysis. We evaluate the averaged point-wise sensitivity at a 40~ms tolerance, in line with recent delineation studies~\cite{chen2023_tol40,chuma2026performance}.
For a more detailed point-wise analysis, we additionally report the positive predictive value (PPV) as a precision counterpart to sensitivity, extend the tolerance to 150~ms for both metrics following earlier studies~\cite{Kalyakulina_2019, AAMI_EC57_2020}, and evaluate localization error as the signed time difference between each predicted onset and offset point and its reference annotation.
Full implementations of all metrics are provided in the Supplementary Material (Section~\ref{supple_evaluation_metrics_code_detail}).

For the comparison among learning strategies, we report the mean and standard deviation over runs with different random seeds to account for training stochasticity (e.g., weight initialization, mini-batch ordering). The statistical significance of the comparison is assessed via paired bootstrap and paired Wilcoxon signed-rank tests, with Holm correction over each family of comparisons~\cite{tibshirani1993introduction, wilcoxon1945individual, holm1979simple}. Details are provided in the Supplementary Material (Section~\ref{supple_perfomance_detail}), which also reports a seed-level robustness check of these comparisons.

\section{Experimental Setup}
\label{sec:experiments}
 
Our study proceeds in two stages: (1) developing deep learning delineation models under supervised, semi-supervised, and self-supervised strategies and selecting the best-performing pretraining objective, and (2) benchmarking the selected model against existing delineation tools.

\subsection{Comparing Learning Strategies for ECG Delineation}
\label{sec:exp-strategy}
The first stage compares learning strategies to see whether self-supervised pretraining benefits delineation and how it interacts with the use of unlabeled data during fine-tuning.
Strategies vary along two factors: (1) whether the encoder is initialized by self-supervised pretraining on a large-scale unlabeled dataset from a separate source or trained from scratch; and (2) whether training uses labeled data only (supervised) or additionally exploits an unlabeled set (semi-supervised).
Crossing the two factors lets us isolate the individual effects of self-supervised pretraining and of semi-supervised training by varying each factor in turn.
For pretraining, we select one representative objective from each of the three major families of SSL:
\begin{itemize}
    \item \textbf{MAE}~\cite{he2022masked}, a generative objective that masks a large fraction of the input and trains the encoder to reconstruct it.
    \item \textbf{MoCo}~\cite{chen2021empirical}, a contrastive objective that pulls together augmented views of the same signal and pushes apart views of different signals.
    \item \textbf{MERL}~\cite{liu2024zero}, a multimodal objective that aligns each ECG with its paired textual report in a shared representation space.
\end{itemize}
For semi-supervised learning, we adopt Mean Teacher (MT)~\cite{tarvainen2017mean}, a consistency-regularization method shown to be strong and robust for ECG delineation~\cite{park2025semisegecg}, with a weak-strong augmentation scheme.

\subsection{Benchmarking Against Existing Delineation Tools}
\label{sec:exp-tools}
The second stage benchmarks the deep learning model developed with the best-performing pretraining objective from the first stage, under both supervised and semi-supervised fine-tuning, against existing open-source (rule-based) and commercial delineation tools.
We evaluate four widely used delineation tools, three open-source and one commercial:
\begin{itemize}
\item \textbf{\textsf{NeuroKit2}}~\cite{makowski2021neurokit2}, a Python toolbox for biosignal processing that includes wavelet-based ECG delineation.
\item \textbf{\textsf{ECGdeli}}~\cite{pilia2021ecgdeli}, a MATLAB toolbox providing wavelet-based delineation.
\item \textbf{\textsf{Prominence}}~\cite{Emrich2024}, a Python implementation of peak-prominence delineation.
\item \textbf{\textsf{CalECG}}~\cite{cohen2025validation}, a commercial package used in regulatory and pharmaceutical settings.
\end{itemize}

\begin{table*}[t]
\centering
\caption{Benchmarking results of learning strategies for ECG delineation. Per-dataset entries are mIoU, reported as mean $\pm$ standard deviation over three seeds, where Internal denotes the \textit{QTDB} $+$ \textit{ISP} test split. Under Summary, Avg.\ is the mIoU averaged across the four datasets for each fine-tuning strategy, and Rank is the average rank among the four pretraining strategies, aggregated over fine-tuning and datasets. The best value in each column is shown in \textbf{bold}.}
\label{tab:strategy}
\begin{tabular}{llcccccc}
\toprule
 & & \multicolumn{4}{c}{mIoU (\%) $\uparrow$} & \multicolumn{2}{c}{Summary} \\
\cmidrule(lr){3-6} \cmidrule(lr){7-8}
Pretraining & Fine-tuning & Internal & \textit{LUDB} & \textit{Zhejiang} & \textit{RDB} & Avg. (\%) $\uparrow$ & Rank $\downarrow$\\
\midrule
\multirow{2}{*}{--}
 & Supervised      & $75.2 \pm 0.2$ & $64.8 \pm 0.8$ & $53.3 \pm 2.4$ & $60.5 \pm 0.5$ & $63.5$ & \multirow{2}{*}{$3.50$} \\
 & Semi-supervised & $81.6 \pm 0.2$ & $70.0 \pm 5.9$ & $66.0 \pm 5.3$ & $67.9 \pm 0.9$ & $71.4$ & \\
\midrule
\multirow{2}{*}{\textbf{MAE}$^{*}$}
 & Supervised      & $\mathbf{82.8 \pm 0.3}$ & $71.3 \pm 2.3$ & $\mathbf{75.2 \pm 1.5}$ & $\mathbf{71.9 \pm 1.2}$ & $\mathbf{75.3}$ & \multirow{2}{*}{$\mathbf{1.25}$} \\
 & Semi-supervised & $82.3 \pm 0.5$ & $72.8 \pm 5.3$ & $74.8 \pm 2.1$ & $71.1 \pm 0.4$ & $75.2$ & \\
\midrule
\multirow{2}{*}{\textbf{MoCo}$^{*}$}
 & Supervised      & $82.2 \pm 0.1$ & $69.5 \pm 3.3$ & $66.2 \pm 3.7$ & $71.6 \pm 1.5$ & $72.4$ & \multirow{2}{*}{$1.75$} \\
 & Semi-supervised & $82.5 \pm 0.1$ & $\mathbf{73.3 \pm 3.9}$ & $73.3 \pm 1.0$ & $70.9 \pm 0.7$ & $75.0$ & \\
\midrule
\multirow{2}{*}{\textbf{MERL}}
 & Supervised      & $76.3 \pm 0.2$ & $67.0 \pm 0.7$ & $52.3 \pm 0.7$ & $61.7 \pm 0.3$ & $64.3$ & \multirow{2}{*}{$3.50$} \\
 & Semi-supervised & $81.0 \pm 0.1$ & $61.8 \pm 2.1$ & $62.6 \pm 5.0$ & $69.4 \pm 0.5$ & $68.7$ & \\
\bottomrule
\end{tabular}

\vspace{2pt}
{\footnotesize $^{*}$: significantly better than without pretraining.}
\end{table*}

\subsection{Implementation Details}
\label{sec:impl}

For the delineation tools, we use the default configurations without manual correction.
For \textsf{NeuroKit2}, two wavelet variants are evaluated: the continuous wavelet transform (CWT) and the discrete wavelet transform (DWT). For \textsf{CalECG}, we use the four fiducial points that define the clinical intervals (P onset, QRS onset, QRS offset, and T offset). It is therefore evaluated on these four fiducial points and on the interval-level metrics, but not on the remaining fiducial points assessed for the other tools. The configuration used for each tool is provided in the Supplementary Material (Section~\ref{sec:supple_tool_cfg}).

The deep learning model architecture is a ViT-Tiny~\cite{dosovitskiy2021an,touvron2021training} encoder with a two-layer FCN decoder~\cite{long2015fully}, using a 128-dimensional hidden layer and a dropout rate of $p=0.1$~\cite{srivastava2014dropout}.
We adopt this encoder for its strong performance in ECG analysis and its effectiveness for semi-supervised ECG segmentation~\cite{park2025semisegecg}.
We keep the architecture fixed across all learning strategies, so that any performance difference is attributable to the strategy rather than to architectural factors.

For pretraining, we follow the training schedule from each objective's original work.
For delineation, both supervised and semi-supervised models are trained for 100 epochs with a batch size of 16, using AdamW~\cite{loshchilov2018decoupled} with weight decay 0.05 and a cosine learning rate schedule~\cite{loshchilov2017sgdr} that warms up to 0.001 over 10 epochs before annealing to 0.0001. For semi-supervised learning, an unlabeled batch of equal size is sampled alongside the labeled batch for consistency regularization.
Each model is trained with three random seeds, and we use the run with the median validation mIoU for the tool comparison.
These experiments are implemented in PyTorch 1.11 with Python 3.9 and run on four NVIDIA RTX 4090 GPUs. Full training configurations are provided in the Supplementary Material (Section~\ref{supple_implementation_detail})

\section{Results}
\label{sec:results}

\subsection{Comparison of Learning Strategies}
\label{sec:res-strategy}

Table~\ref{tab:strategy} reports the delineation performance (mIoU) of the different learning strategies across all test sets, with the averaged interval error and the averaged point-wise sensitivity results provided in the Supplementary Material (Tables~\ref{tab:interval}--\ref{tab:loc-strategy-pointwise}).

\textbf{Without pretraining, semi-supervised learning improves over supervised learning.}
The improvement is largely consistent across metrics and datasets.
mIoU improves on all datasets (63.5 to 71.4\% on average), and averaged point-wise sensitivity shows a similar trend.
The averaged interval error also improves on all datasets except \textit{RDB}, where supervised learning gives a lower error.
\textit{Zhejiang} shows the largest gains across all three metrics, with mIoU, for example, rising from 53.3 to 66.0\%.
Given that \textit{Zhejiang} consists entirely of recordings with arrhythmic beats, this suggests that the benefit is pronounced under arrhythmic conditions.

\textbf{Self-supervised pretraining helps, but the objective matters.}
Across all datasets, the best performance is attained by a pretrained strategy. The benefit of pretraining, however, depends on the objective. \textbf{MAE} and \textbf{MoCo} significantly outperform no pretraining, whereas \textbf{MERL} does not. Averaged across datasets, \textbf{MERL} underperforms the other objectives by a clear margin (mIoU 68.7 vs. 75.2\% for \textbf{MAE} and 75.0\% for \textbf{MoCo} under semi-supervised fine-tuning). On the internal test set, which is derived from the same data source as the training data, \textbf{MAE} and \textbf{MoCo} are near saturation (mIoU 82.2--82.8\%) and even \textbf{MERL} approaches them under semi-supervised fine-tuning (81.0 vs.\ 82.3--82.5\%). These differences among the objectives widen on the external sets. Furthermore, the two effective objectives interact differently with fine-tuning. \textbf{MAE} performs comparably under supervised and semi-supervised fine-tuning (mIoU $75.3$ vs.\ $75.2$\%), whereas \textbf{MoCo} benefits substantially from semi-supervised fine-tuning (mIoU $72.4$ vs.\ $75.0$\%), most clearly on \textit{Zhejiang} (mIoU $66.2\rightarrow73.3$\%), with gains of similar magnitude in averaged interval error and averaged point-wise sensitivity.

\textbf{MAE} and \textbf{MoCo}, the two effective objectives, show no significant difference. We adopt \textbf{MAE} for its higher mIoU on average and better average rank, as well as its stability under both supervised and semi-supervised fine-tuning, which \textbf{MoCo} lacks.
The resulting models under supervised and semi-supervised fine-tuning, denoted \textsc{DL delineator} (sup) and (semi), serve as the deep learning models in the analyses that follow.

\begin{table*}[t]
\centering
\scriptsize
\setlength{\tabcolsep}{4pt}
\caption{Comparison of the DL delineator with open-source delineation tools. The best value in each column is in \textbf{bold}, and the second-best is \underline{underlined}. Avg.\ Rank is the mean rank across all columns. \textsf{CalECG}, configured for interval outputs only, is compared separately on interval error in the Supplementary Material (Table~\ref{tab:calecg_interval}).}
\label{tab:tools_main}
\resizebox{\textwidth}{!}{%
\begin{tabular}{@{}l cccc cccc cccc c@{}}
\toprule
& \multicolumn{4}{c}{mIoU (\%) $\uparrow$}
& \multicolumn{4}{c}{Averaged interval error (ms) $\downarrow$}
& \multicolumn{4}{c}{Averaged point-wise sensitivity (\%) $\uparrow$}
& \\
\cmidrule(lr){2-5}\cmidrule(lr){6-9}\cmidrule(lr){10-13}
Method & Internal & \textit{LUDB} & \textit{Zhejiang} & \textit{RDB}
       & Internal & \textit{LUDB} & \textit{Zhejiang} & \textit{RDB}
       & Internal & \textit{LUDB} & \textit{Zhejiang} & \textit{RDB}
       & Avg.\ Rank $\downarrow$ \\
\midrule
\textsf{NeuroKit2} (CWT)
 & 41.3 & 44.9 & 7.3 & 40.0
 & 70.9 & 108.3 & 72.8 & 30.2
 & 59.4& 59.9 & 15.7 & 54.9& 5.8\\
\textsf{NeuroKit2} (DWT)
 & 43.4 & 45.6 & 8.6 & 38.6
 & 43.6 & 45.6 & 52.7 & 40.5
 & 60.3& 62.2 & 17.8 & 58.5& 5.2\\
\textsf{ECGdeli}
 & 67.5 & 59.4 & 41.9 & 54.8
 & 24.6 & 29.3 & 30.4 & 30.0
 & 82.4 & 80.6 & 56.3 & 76.4
 & 3.4\\
\textsf{Prominence}
 & 60.4 & 63.9 & 42.7 & 51.7
 & 28.1 & 22.3 & 35.9 & 23.0
 & 78.1 & 82.9 & 54.6 & 73.7
 & 3.6\\
\midrule
\textsc{DL delineator} (sup)
 & \textbf{83.0}& \underline{71.6} & \underline{73.5} & \underline{71.1}
 & \textbf{11.5} & \underline{19.0} & \underline{16.3} & \underline{22.2}
 & \textbf{96.1} & \underline{86.6} & \underline{82.9} & \textbf{92.6}
 & \underline{1.8}\\
\textsc{DL delineator} (semi)
 & \underline{82.4}& \textbf{75.8} & \textbf{75.1} & \textbf{71.3}
 & \underline{12.0} & \textbf{16.8} & \textbf{15.3} & \textbf{21.7}
 & \textbf{96.1} & \textbf{91.1} & \textbf{86.0} & \underline{92.5}
 & \textbf{1.3}\\
\bottomrule
\end{tabular}%
}
\end{table*}

\subsection{Comparison with Existing Tools}
\label{sec:res-tools}
Table~\ref{tab:tools_main} reports the benchmark against existing delineation tools across the internal and three external test sets (\textit{LUDB}, \textit{Zhejiang}, and \textit{RDB}). Among the existing tools, \textsf{ECGdeli} and \textsf{Prominence} are the top two, with average ranks of 3.4 and 3.6, respectively. The \textsc{DL delineator} significantly surpasses all the benchmarked tools across every metric and dataset, with its two fine-tuning configurations taking the best and second-best average ranks (1.3 and 1.7, respectively).

The advantage of the \textsc{DL delineator} is most pronounced on datasets with arrhythmias. On \textit{RDB}, which spans a wide range of rhythms, the \textsc{DL delineator} clearly outperforms the strongest tool, \textsf{ECGdeli} (mIoU 71.3 vs. 54.8\%; averaged point-wise sensitivity 92.6 vs. 76.4\%). The margin is largest on \textit{Zhejiang} (mIoU 75.1 vs. 42.7\% for \textsf{Prominence}; averaged interval error 15.3 vs. 30.4 ms and averaged point-wise sensitivity 86.0 vs. 56.3\% for \textsf{ECGdeli}). 
For the point-wise sensitivity, these margins narrow under the less stringent 150~ms tolerance, but their ordering is unchanged; the same pattern holds for the point-wise PPV in the Supplementary Material (Table~\ref{tab:tol_supp_sens_ppv}).

Table~\ref{tab:tools_mecgdb} reports the averaged interval error on \textit{mECGDB}. The \textsc{DL delineator} again achieves the lowest error, reducing the average from 25.4 ms for the best tool (\textsf{Prominence}) to 15.0 ms.
Both configurations improve significantly on the average, the PR interval, and the QRS duration.
For the QT interval, both remain the most accurate, each with an error of 19.1~ms compared with 21.7~ms for \textsf{Prominence}, and the improvement is significant in all cases except the supervised configuration against \textsf{Prominence}.
Its advantage therefore holds even for portable, out-of-clinic recordings.

\begin{table}[t]
\centering
\footnotesize
\setlength{\tabcolsep}{6pt}
\caption{Comparison of the DL delineator on \textit{mECGDB} with open-source and commercial delineation tools, evaluated by mean absolute error (ms) for the PR interval, QRS duration, QT interval, and their average. The best value is in \textbf{bold} and the second-best is \underline{underlined}.}
\label{tab:tools_mecgdb}
\begin{tabular}{@{}l c|ccc@{}}
\toprule
Method & Average & PR & QRS & QT \\
\midrule
\textsf{NeuroKit2} (CWT) & 97.3 & 207.4 & 32.1 & 52.3 \\
\textsf{NeuroKit2} (DWT) & 51.3 & 45.3 & 17.0 & 91.5 \\
\textsf{ECGdeli}         & 32.9 & 47.0 & 18.0 & 33.6 \\
\textsf{CalECG}          & 37.9 & 41.1 & 32.7 & 40.0 \\
\textsf{Prominence}      & 25.4 & 29.6 & 25.0 & 21.7 \\
\midrule
\textsc{DL delineator} (sup)    & \textbf{15.0} & \textbf{13.8} & \underline{12.2} & \textbf{19.1} \\
\textsc{DL delineator} (semi)    & \underline{15.4} & \underline{15.3} & \textbf{11.8} & \textbf{19.1} \\
\bottomrule
\end{tabular}
\end{table}

\begin{table*}[t]
\centering
\footnotesize
\setlength{\tabcolsep}{5pt}
\caption{Comparison of the DL delineator on external test sets (\textit{LUDB} and \textit{Zhejiang}) with open-source and commercial delineation tools, evaluated by point-wise sensitivity (\%) per fiducial point. "Total" is the average over the available onset/offset points. The best value is in \textbf{bold} and the second-best is \underline{underlined}.}
\label{tab:pointwise_se}
\begin{tabular}{@{}ll ccccccc@{}}
\toprule
Dataset & Method & P$_{\mathrm{on}}$ & P$_{\mathrm{off}}$ & QRS$_{\mathrm{on}}$ & QRS$_{\mathrm{off}}$ & T$_{\mathrm{on}}$ & T$_{\mathrm{off}}$ & Total \\
\midrule
\multirow{6}{*}{\textit{LUDB}}
 & \textsf{NeuroKit2} (CWT) & 43.7 & 43.1 & 80.8 & 72.2 & 56.4 & 63.4 & 59.9 \\
 & \textsf{NeuroKit2} (DWT) & 73.3 & 63.8 & 66.2 & 81.8 & 32.5 & 55.8 & 62.2 \\
 & \textsf{ECGdeli}         & 71.8 & \underline{87.2} & 97.3 & 85.1 & 70.7 & 71.7 & 80.6 \\
 & \textsf{CalECG}          & 74.0 & - & 67.7 & 69.1 & - & 61.5 & 68.1 \\
 & \textsf{Prominence}      & 79.2 & 80.8 & 91.9 & 85.8 & \underline{79.1} & 80.7 & 82.9 \\
 \cmidrule(lr){2-9}
 & \textsc{DL delineator} (sup)   & \underline{80.4} & 81.3 & \underline{99.1} & \underline{98.9} & 77.8 & \underline{81.6} & 
 \underline{86.6} \\
 & \textsc{DL delineator} (semi)  & \textbf{94.0} & \textbf{90.3} & \textbf{99.4} & \textbf{99.2} & \textbf{79.2} & \textbf{84.6} & 
 \textbf{91.2} \\
\midrule
\multirow{6}{*}{\textit{Zhejiang}}
 & \textsf{NeuroKit2} (CWT) & 14.7 & 13.6 & 20.8 & 24.5 & 9.8 & 11.0 & 15.7 \\
 & \textsf{NeuroKit2} (DWT) & 24.4 & 24.6 & 14.3 & 20.0 & 11.4 & 11.8 & 17.8 \\
 & \textsf{ECGdeli}         & 57.7 & 37.3 & 83.2 & 67.4 & 46.5 & 45.4 & 56.3 \\
 & \textsf{CalECG}          & 5.6 & - & 5.7 & 9.6 & - & 7.2 & 7.0 \\
 & \textsf{Prominence}      & 65.4 & 64.8 & 68.0 & 66.4 & 32.5 & 30.4 & 54.6 \\
 \cmidrule(lr){2-9}
 & \textsc{DL delineator} (sup) & \underline{82.6} & \underline{79.9} & \underline{92.0} & \underline{93.2} & \underline{69.1} & \underline{80.9} & \underline{82.9} \\
 & \textsc{DL delineator} (semi) & \textbf{87.5} & \textbf{84.1} & \textbf{94.6} & \textbf{94.6} & \textbf{69.6} & \textbf{85.4} & \textbf{86.0} \\
\bottomrule
\end{tabular}
\end{table*}

\subsection{Point-wise Performance Characteristics}
\label{sec:res-toolwise}

To characterize each tool's distinctive behavior, we examine point-wise sensitivity on \textit{LUDB} and \textit{Zhejiang} (Table~\ref{tab:pointwise_se}), chosen for their contrasting morphology: \textit{LUDB} is dominated by sinus rhythms, whereas \textit{Zhejiang} consists entirely of recordings with ventricular ectopic and tachycardic beats. The corresponding point-wise PPV and localization errors (signed mean $\pm$ SD) are reported in the Supplementary Material (Table~\ref{tab:pointwise_ppv}--\ref{tab:pointwise_error}).

The two \textsf{NeuroKit2} variants are each limited at a different fiducial point of the beat. 
The CWT-based method shows particularly low sensitivity at P\textsubscript{on} (43.7 and 14.7\% on \textit{LUDB} and \textit{Zhejiang}, respectively), whereas the DWT-based method struggles at QRS\textsubscript{on} and T\textsubscript{off}, the two fiducial points that define the QT interval (e.g., 66.2 and 55.8\% on \textit{LUDB}). These observations are consistent with their errors on \textit{mECGDB}, where the CWT-based and DWT-based methods show large PR and QT interval errors, respectively (207.4 and 91.5~ms; Table~\ref{tab:tools_mecgdb}).

\textsf{CalECG} was configured to output only the fiducial points required to define the clinical intervals; it therefore provides P\textsubscript{on}, QRS\textsubscript{on}, QRS\textsubscript{off}, and T\textsubscript{off}, but not P\textsubscript{off} or T\textsubscript{on}, and was evaluated only at these four points. Among them, its sensitivity was moderate on \textit{LUDB} (68.1\% overall) but well below \textsf{ECGdeli} and \textsf{Prominence}, and it dropped sharply under arrhythmia, reaching only 7.0\% overall on \textit{Zhejiang}.

\textsf{ECGdeli} and \textsf{Prominence} achieve the highest performance among the evaluated tools, although their superiority varies across datasets and fiducial points. \textsf{ECGdeli} is more accurate on the QRS complex, giving the highest QRS\textsubscript{on} sensitivity among the tools on both \textit{LUDB} and \textit{Zhejiang} (97.3 and 83.2\%, respectively), whereas \textsf{Prominence} is more reliable at P\textsubscript{on} (79.2 and 65.4\%, respectively). For the remaining points, the leading tool depends on the dataset.

By contrast, the \textsc{DL delineator} surpasses every tool at nearly all fiducial points. The two fine-tuning configurations differ only at a few points on \textit{LUDB}: supervised fine-tuning falls short of the best tool at P\textsubscript{off} (81.3 vs. 87.2\% for \textsf{ECGdeli}) and T\textsubscript{on} (77.8 vs. 79.1\% for \textsf{Prominence}), whereas semi-supervised fine-tuning exceeds all tools at both (90.3 and 79.2\%). On \textit{Zhejiang}, both configurations lead at every point by a wide margin (e.g., T\textsubscript{off} 85.4 vs. 45.4\% for \textsf{ECGdeli}).

\begin{table*}[t]
\centering
\footnotesize
\setlength{\tabcolsep}{6pt}
\caption{Rhythm-stratified subgroup performance on \textit{RDB} for the sinus and arrhythmia groups, with the relative performance degradation from sinus to arrhythmia (smaller is better). The best value is in \textbf{bold} and the second-best is \underline{underlined}.}
\label{tab:subgroup_main}
\begin{tabular}{@{}llccc@{}}
\toprule
& & \multicolumn{2}{c}{Subgroup performance} & \\
\cmidrule{3-4}
Metric & Method & Sinus & Arrhythmia & Degradation \\
\midrule
\multirow{4}{*}{mIoU (\%) $\uparrow$}
 & \textsf{ECGdeli}               & 66.3 & 38.7 & 41.6\% \\
 & \textsf{Prominence}            & 64.4 & 33.2 & 48.5\% \\
 & \textsc{DL delineator} (sup)   & \underline{76.1} & \textbf{64.9} & \textbf{14.7\%} \\
 & \textsc{DL delineator} (semi)  & \textbf{76.6} & \underline{64.2} & \underline{16.2\%} \\
\midrule
\multirow{4}{*}{Averaged interval error (ms) $\downarrow$}
 & \textsf{ECGdeli}               & 28.9 & 40.1 & \underline{38.8\%} \\
 & \textsf{Prominence}            & \textbf{18.2} & 38.2 & 110.0\% \\
 & \textsc{DL delineator} (sup)   & 20.6 & \textbf{28.5} & \textbf{38.4\%} \\
 & \textsc{DL delineator} (semi)  & \underline{19.4} & \underline{30.2} & 56.0\% \\
\midrule
\multirow{4}{*}{Averaged point-wise sensitivity (\%) $\uparrow$}
 & \textsf{ECGdeli}               & 81.1 & 67.3 & 17.1\% \\
 & \textsf{Prominence}            & 83.5 & 58.3 & 30.2\% \\
 & \textsc{DL delineator} (sup)   & \underline{96.4} & \underline{80.1} & \underline{16.9\%} \\
 & \textsc{DL delineator} (semi)  & \textbf{96.5} & \textbf{80.4} & \textbf{16.7\%} \\
\bottomrule
\end{tabular}
\end{table*}

\subsection{Rhythm-stratified Subgroup Analysis}
\label{sec:res-subgroup}
\textit{RDB} contains a large and diverse set of rhythms, making it well-suited for a rhythm-stratified subgroup analysis. We evaluate all metrics for each rhythm and compare the \textsc{DL delineator} in detail with the two leading tools, \textsf{ECGdeli} and \textsf{Prominence}. In Table~\ref{tab:subgroup_main}, rhythms are grouped into sinus and arrhythmia; we report each group's mean and the relative degradation between them, with per-rhythm results in the Supplementary Material (Table~\ref{tab:subgroup_supp}).

In the sinus group, both configurations of the \textsc{DL delineator} lead in mIoU and averaged point-wise sensitivity by a wide margin (e.g., mIoU 76.1 and 76.6 vs. 66.3\% for \textsf{ECGdeli}), and trail only \textsf{Prominence} on averaged interval error, by a small margin (19.4 vs.\ 18.2 ms).
Under arrhythmia, both configurations are the most accurate on every metric, outperforming the best tool on each (mIoU 64.9 vs. 38.7\%; interval error 28.5 vs.\ 38.2 ms; sensitivity 80.4 vs. 67.3\%).

The \textsc{DL delineator} is also the most robust to sinus-to-arrhythmia shift.
On mIoU, its relative degradation is far smaller than that of either tool (14.7 and 16.2 vs.\ 41.6 and 48.5\%), and on sensitivity it is again the smallest (16.7 vs.\ 17.1\% for \textsf{ECGdeli}).
Notably, the \textsc{DL delineator} combines strong accuracy on sinus rhythms with the smallest relative degradation, maintaining its lead under arrhythmia.
For interval error, \textsf{Prominence} degrades sharply (110.0\%) and \textsf{ECGdeli} starts from the worst error in both groups (28.9 and 40.1~ms), whereas the \textsc{DL delineator} keeps the lowest interval error under arrhythmia.

\begin{figure*}[t]
    \centering
    \includegraphics[width=1.0\linewidth]{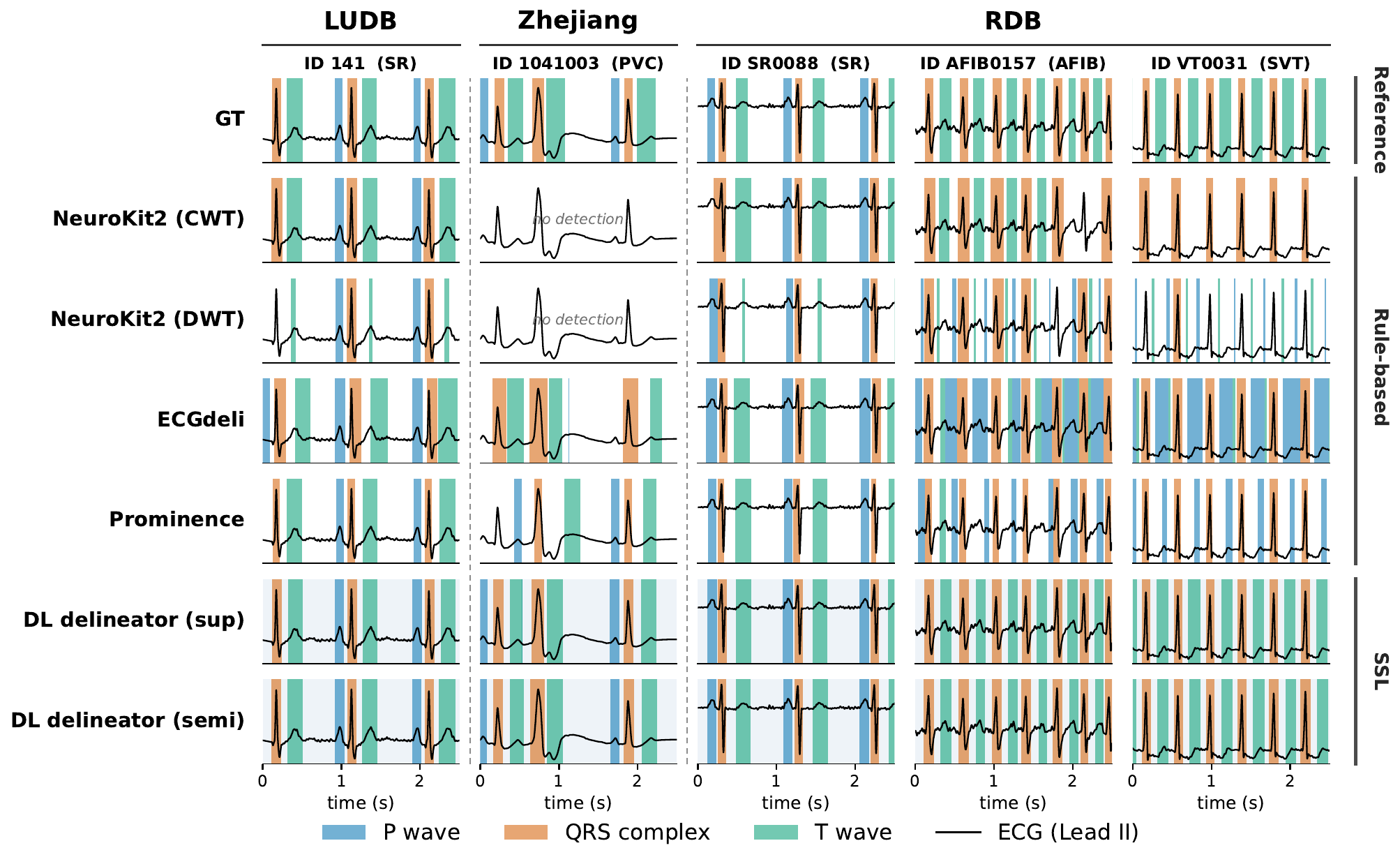}
    \caption{Qualitative examples of ECG delineation results across methods and datasets. The rhythm classes shown reflect the characteristics of each dataset. Rows are grouped into the reference annotation (GT), rule-based tools, and the \textsc{DL delineator}; shaded bands mark each method's detected P (blue), QRS (orange), and T (green) waves on the \emph{same} input signal. SR, sinus rhythm; PVC, premature ventricular contraction; AFIB, atrial fibrillation; SVT, supraventricular tachycardia. }
    \label{fig:main_quality}
\end{figure*}

\subsection{Qualitative Analysis}
Fig.~\ref{fig:main_quality} shows representative delineation results across methods and datasets, with additional per-dataset examples in the Supplementary Material (Fig.~\ref{fig:quality_ludb_3samples}--\ref{fig:quality_rdb_3samples}), which follow similar trends. The examples make the differences between methods visually apparent and align with the quantitative results.

On sinus rhythms (\textit{LUDB} and \textit{RDB}), most methods recover the overall P, QRS, and T structure, and the differences appear mainly at the P and QRS onsets and offsets. \textsf{NeuroKit2} is the exception, remaining unreliable even on sinus rhythms: the CWT-based method misses the first P wave on \textit{RDB}, and the DWT-based method misses the first QRS on \textit{LUDB} and delineates the T wave too narrowly. These failures are consistent with the low point-wise sensitivity reported earlier, at P\textsubscript{on} for the CWT-based method and at the QRS and T boundaries for the DWT-based method.

The differences become pronounced under arrhythmia. On \textit{Zhejiang}, which contains premature ventricular complexes, both \textsf{NeuroKit2} variants fail to delineate; \textsf{ECGdeli} misses the P wave; and \textsf{Prominence} misses the P wave and QRS complex of the beat preceding the ectopic beat, whereas the \textsc{DL delineator} stays close to the reference. On \textit{RDB}, the existing tools produce characteristic false positives under non-sinus rhythm. For example, under atrial fibrillation and supraventricular tachycardia, both \textsf{ECGdeli} and \textsf{Prominence} mark spurious P waves, in the latter case by labeling T waves as P waves. In contrast, both configurations of the \textsc{DL delineator} remain aligned with the reference across all rhythm types.

Overall, these qualitative results mirror the point-wise and rhythm-stratified findings: the existing tools are reliable on sinus rhythms but lose accuracy at the P and T boundaries and break down under arrhythmia, whereas the \textsc{DL delineator} stays robust across datasets and rhythm types.

\section{Discussion}
\label{sec:discussion}
 
\subsection{Effectiveness of Label-Efficient Deep Learning}
\label{sec:disc-ssl}
Self-supervised pretraining is effective for ECG delineation. Pretraining on a large unlabeled dataset with a suitable objective improves performance across all test sets, with the clearest gains on the external sets (Table~\ref{tab:strategy}). A plausible interpretation is that unlabeled data exposes the model to a wider range of morphologies, rhythms, and noise than the labeled set alone, which should matter most when the test distribution deviates from that of the labeled training set. These results establish self-supervised pretraining as a practical approach to label-efficient delineation and broaden the strategies available for this task.

The choice of pretraining objective also matters. The reconstruction objective (\textbf{MAE}) and the unimodal contrastive objective (\textbf{MoCo}) transfer well, whereas the multimodal contrastive objective (\textbf{MERL}), which additionally uses paired text reports, underperformed. Because \textbf{MoCo} is also contrastive, the gap does not appear to stem from contrastive learning itself; one possibility is that aligning ECGs to text encourages invariance to the fine morphological detail that delineation depends on. The benefit of self-supervised pretraining therefore depends on the design of the pretraining objective, not on self-supervised learning alone, making an objective tailored to per-time-point localization a promising direction for future work.

Self-supervised pretraining and semi-supervised learning each improve delineation over supervised learning, but their benefits do not simply add up (Table~\ref{tab:strategy}). The contribution of semi-supervised learning depends on how much room the supervised baseline leaves: it is the largest without pretraining, remains substantial for \textbf{MoCo}, and is negligible for \textbf{MAE}, whose supervised fine-tuning performance is already saturated. A plausible interpretation is that a sufficiently effective pretraining objective already captures much of what the unlabeled recordings offer, leaving little for semi-supervised fine-tuning to add. Because it is difficult to know in advance whether a given objective leaves such room, semi-supervised fine-tuning is a safe default across the objectives we examine: it helps when room remains and at least does not degrade performance.

\subsection{The Advantage of \textsc{DL delineator} over Existing Tools}
\label{sec:disc-tools}
Across the internal and three external sets, the \textsc{DL delineator} outperforms every tool on all three metrics, and its improvements are statistically significant in all comparisons (Table~\ref{tab:tools_main}). On a further external set recorded outside clinical conditions, \textit{mECGDB}, it again achieves the lowest error (Table~\ref{tab:tools_mecgdb}). The advantage holds under both fine-tuning strategies, consistent with our choice of \textbf{MAE}, which we select for its stability across fine-tuning schemes. Together, these results provide systematic evidence that label-efficient deep learning offers a measurable and consistent advantage over existing open-source and commercial tools, addressing whether it is worth adopting for delineation in practice.

\subsection{Performance across Fiducial Points and Rhythms}
A point-wise analysis (Table~\ref{tab:pointwise_se}) localizes where the two fine-tuning strategies differ: semi-supervised fine-tuning closes the residual gaps that supervised fine-tuning leaves at P\textsubscript{off} and T\textsubscript{on} on \textit{LUDB}. Because this comparison uses a single selected model per configuration, we read these differences as tendencies rather than precise magnitudes. This refines, rather than contradicts, the saturation reported in Section~\ref{sec:disc-ssl}: that saturation is in the averaged metric, whereas these differences are confined to individual boundaries. Even when the averaged metric shows no measurable improvement, semi-supervised fine-tuning thus improves a few boundaries without degrading others, supporting its use as a low-risk default. Beyond this difference, the \textsc{DL delineator} (semi) is consistent across fiducial points; its weakest region is the T-wave boundaries, and T\textsubscript{on} in particular, where the sensitivity is the lowest of the six points. These are the hardest points for every tool, reflecting the gradual, low-amplitude onset of the T-wave rather than a limitation specific to our model, so improving localization there while maintaining overall performance is a worthwhile direction for future work.

We next stratify the same results by rhythm (Table~\ref{tab:subgroup_main}), a property of direct practical importance, since abnormal rhythms are common in clinical recordings and are harder to delineate. The \textsc{DL delineator} retains its accuracy under arrhythmia, remaining the most accurate method on every metric. This is consistent with learned representations capturing morphological variability that fixed heuristics cannot. Moreover, rule-based tools are often evaluated on predominantly sinus recordings, so their reported accuracy may not carry over to the arrhythmic conditions common in practice.

\subsection{From Tool Selection to a Single Robust Delineator}
Among the tools, \textsf{ECGdeli} and \textsf{Prominence} are the strongest and essentially indistinguishable in average rank (3.5 vs.\ 3.7). For users who rely on them, the choice is better framed by what is measured and on which rhythm than by which tool is better overall. On predominantly sinus recordings, \textsf{Prominence} is a reasonable default: it leads \textsf{ECGdeli} across metrics on \textit{LUDB} and in the sinus subgroup of \textit{RDB}, and is more reliable at P\textsubscript{on}, making it preferable when P-onset localization is required (Tables~\ref{tab:tools_main},~\ref{tab:pointwise_se},~\ref{tab:subgroup_main}). \textsf{ECGdeli} is preferable when the QRS complex is the target (at QRS\textsubscript{on} and QRS duration) and is the safer choice when rhythm composition is mixed or unknown, degrading less from sinus to arrhythmia and holding up better on the T wave under arrhythmia (Tables~\ref{tab:tools_mecgdb},~\ref{tab:pointwise_se},~\ref{tab:subgroup_main}).

In practice, however, such tool selection is hard to apply. Under arrhythmia, both tools degraded sharply (Table~\ref{tab:subgroup_main}), and choosing the right tool presupposes knowing the rhythm of each incoming recording in advance, which requires an additional classification step. A single learned delineator avoids this: the \textsc{DL delineator} remains robust across acquisition settings and rhythm types, so it can be deployed without characterizing each recording or switching tools between settings, while remaining more accurate than either tool. Moreover, it runs faster than these tools (Supplementary Material, Table~\ref{tab:latency}), so its accuracy does not come at the cost of speed.

\subsection{Limitations and Future Work}
\label{sec:disc-limitations} 
Our study has several limitations. First, we examined only a subset of possible self-supervised approaches (three objectives with a single, established backbone), so the strategies we identify are effective, practical directions rather than the optimal choice. A broader exploration of self-supervised objectives is therefore a natural next step. Because the benefit of pretraining depends strongly on the objective, with \textbf{MAE} (a dense reconstruction task) transferring best, a promising direction is to move beyond generic reconstruction toward pretext tasks that target temporal localization more directly, such as objectives built around the wave boundaries that delineation must predict. Second, \textit{mECGDB} provides interval-level labels alone, so the evidence under uncontrolled conditions rests on a single metric; recordings annotated across a wider range of devices would allow the full metric set to be assessed under such conditions.

\section{Conclusion}
\label{sec:conclusion}
We studied label-efficient deep learning for ECG delineation and compared it, under identical conditions, against widely used open-source and commercial tools on one internal and four external test sets using three complementary metrics. Self-supervised pretraining is effective, though its benefit depends on the objective: with \textbf{MAE} pretraining, the resulting \textsc{DL delineator} outperforms every tool on all three metrics across datasets under both fine-tuning configurations, with the improvement statistically significant in nearly all comparisons and its margin widening under arrhythmia. Resolving performance by fiducial point and rhythm further shows that the tools behave differently, concentrating at the P and T waves and under arrhythmia, whereas the \textsc{DL delineator} remains robust throughout. Overall, self-supervised pretraining enables deep learning models to surpass existing delineation tools using only a modest amount of labeled data. This work supports wider adoption of label-efficient learning and lays the groundwork for reliable ECG delineation in clinical practice. 

\section*{Data Availability}
All datasets except \textit{mECGDB} are publicly available. For \textit{QTDB}, \textit{ISP}, \textit{PTB-XL}, \textit{LUDB}, and \textit{Zhejiang}, we used the preprocessed versions provided by the \textsf{SemiSegECG} benchmark\footnote{\url{https://github.com/vuno/semi-seg-ecg}}. \textit{RDB}\footnote{\url{https://doi.org/10.6084/m9.figshare.28892186.v3}} and \textit{MIMIC-IV-ECG}\footnote{\url{https://doi.org/10.13026/4nqg-sb35}} are available from their original sources.
\textit{mECGDB} is a private dataset and cannot be shared due to user privacy and ethical restrictions.
However, it can be requested from the corresponding author, subject to institutional approval and user consent.

\section*{References}

\bibliographystyle{IEEEtran}

\bibliography{main}

@book{gacek2011ecg,
  author={Gacek, Adam and Pedrycz, Witold},
  title={ECG signal processing, classification and interpretation: a comprehensive framework of computational intelligence},
  year={2011},
  publisher={Springer Science \& Business Media}
}

@article{rautaharju2009aha,
  title={AHA/ACCF/HRS recommendations for the standardization and interpretation of the electrocardiogram: part IV: the ST segment, T and U waves, and the QT interval: a scientific statement from the American Heart Association Electrocardiography and Arrhythmias Committee, Council on Clinical Cardiology; the American College of Cardiology Foundation; and the Heart Rhythm Society: endorsed by the International Society for Computerized Electrocardiology},
  author={Rautaharju, Pentti M and Surawicz, Borys and Gettes, Leonard S},
  journal={Circulation},
  volume={119},
  number={10},
  pages={e241--e250},
  year={2009},
  publisher={Lippincott Williams \& Wilkins}
}

@article{kligfield2007recommendations,
  title={Recommendations for the standardization and interpretation of the electrocardiogram: part I: the electrocardiogram and its technology: a scientific statement from the American Heart Association Electrocardiography and Arrhythmias Committee, Council on Clinical Cardiology; the American College of Cardiology Foundation; and the Heart Rhythm Society endorsed by the International Society for Computerized Electrocardiology},
  author={Kligfield, Paul and Gettes, Leonard S and Bailey, James J and Childers, Rory and Deal, Barbara J and Hancock, E William and Van Herpen, Gerard and Kors, Jan A and Macfarlane, Peter and Mirvis, David M and others},
  journal={Circulation},
  volume={115},
  number={10},
  pages={1306--1324},
  year={2007},
  publisher={Lippincott Williams \& Wilkins}
}

@article{elgendi2014revisiting,
  author={Elgendi, Mohamed and Eskofier, Bj{\"o}rn and Dokos, Socrates and Abbott, Derek},
  title={Revisiting QRS detection methodologies for portable, wearable, battery-operated, and wireless ECG systems},
  journal={PloS one},
  volume={9},
  number={1},
  pages={e84018},
  year={2014},
  publisher={Public Library of Science San Francisco, USA}
}

@article{liu2018performance,
  author={Liu, Feifei and Liu, Chengyu and Jiang, Xinge and Zhang, Zhimin and Zhang, Yatao and Li, Jianqing and Wei, Shoushui},
  title={Performance analysis of ten common QRS detectors on different ECG application cases},
  journal={Journal of healthcare engineering},
  volume={2018},
  number={1},
  pages={9050812},
  year={2018},
  publisher={Wiley Online Library}
}

@inproceedings{jimenez2019u,
  title={U-Net architecture for the automatic detection and delineation of the electrocardiogram},
  author={Jimenez-Perez, Guillermo and Alcaine, Alejandro and Camara, Oscar},
  booktitle={2019 Computing in Cardiology (CinC)},
  pages={Page--1},
  year={2019},
  organization={IEEE}
}

@article{liang2022ecg_segnet,
  title={ECG\_SegNet: An ECG delineation model based on the encoder-decoder structure},
  author={Liang, Xiaohong and Li, Liping and Liu, Yuanyuan and Chen, Dan and Wang, Xinpei and Hu, Shunbo and Wang, Jikuo and Zhang, Huan and Sun, Chengfa and Liu, Changchun},
  journal={Computers in biology and medicine},
  volume={145},
  pages={105445},
  year={2022},
  publisher={Elsevier}
}

@article{joung2024deep,
  title={Deep learning based ECG segmentation for delineation of diverse arrhythmias},
  author={Joung, Chankyu and Kim, Mijin and Paik, Taejin and Kong, Seong-Ho and Oh, Seung-Young and Jeon, Won Kyeong and Jeon, Jae-hu and Hong, Joong-Sik and Kim, Wan-Joong and Kook, Woong and others},
  journal={PloS one},
  volume={19},
  number={6},
  pages={e0303178},
  year={2024},
  publisher={Public Library of Science San Francisco, CA USA}
}

@article{jin2023label,
  title={Label-efficient deep learning in medical image analysis: Challenges and future directions},
  author={Jin, Cheng and Guo, Zhengrui and Lin, Yi and Luo, Luyang and Chen, Hao},
  journal={arXiv preprint arXiv:2303.12484},
  year={2023}
}

@article{pelaez2023survey,
  title={A survey on semi-supervised semantic segmentation},
  author={Pel{\'a}ez-Vegas, Adrian and Mesejo, Pablo and Luengo, Juli{\'a}n},
  journal={arXiv preprint arXiv:2302.09899},
  year={2023}
}

@article{krishnan2022self,
  title={Self-supervised learning in medicine and healthcare},
  author={Krishnan, Rayan and Rajpurkar, Pranav and Topol, Eric J},
  journal={Nature Biomedical Engineering},
  volume={6},
  number={12},
  pages={1346--1352},
  year={2022},
  publisher={Nature Publishing Group UK London}
}

@article{goldberger2000physiobank,
  title={PhysioBank, PhysioToolkit, and PhysioNet: components of a new research resource for complex physiologic signals},
  author={Goldberger, Ary L and Amaral, Luis AN and Glass, Leon and Hausdorff, Jeffrey M and Ivanov, Plamen Ch and Mark, Roger G and Mietus, Joseph E and Moody, George B and Peng, Chung-Kang and Stanley, H Eugene},
  journal={circulation},
  volume={101},
  number={23},
  pages={e215--e220},
  year={2000},
  publisher={Lippincott Williams \& Wilkins}
}

@inproceedings{kiyasseh2021clocs,
  title={Clocs: Contrastive learning of cardiac signals across space, time, and patients},
  author={Kiyasseh, Dani and Zhu, Tingting and Clifton, David A},
  booktitle={International conference on machine learning},
  pages={5606--5615},
  year={2021},
  organization={PmLR}
}

@inproceedings{na2024guiding,
  title={Guiding masked representation learning to capture spatio-temporal relationship of electrocardiogram},
  author={Na, Yeongyeon and Park, Minje and Tae, Yunwon and Joo, Sunghoon},
  booktitle={International conference on learning representations},
  volume={2024},
  pages={15012--15035},
  year={2024}
}

@article{liu2024zero,
  title={Zero-shot ecg classification with multimodal learning and test-time clinical knowledge enhancement},
  author={Liu, Che and Wan, Zhongwei and Ouyang, Cheng and Shah, Anand and Bai, Wenjia and Arcucci, Rossella},
  journal={arXiv preprint arXiv:2403.06659},
  year={2024}
}

@inproceedings{al2026benchmarking,
  title={Benchmarking ECG FMs: A reality check across clinical tasks},
  author={Al-Masud, MA and Lopez Alcaraz, Juan Miguel and Strodthoff, Nils},
  booktitle={International Conference on Learning Representations},
  volume={2026},
  pages={146689--146715},
  year={2026}
}

@article{martinez2004wavelet,
  title={A wavelet-based ECG delineator: evaluation on standard databases},
  author={Mart{\'\i}nez, Juan Pablo and Almeida, Rute and Olmos, Salvador and Rocha, Ana Paula and Laguna, Pablo},
  journal={IEEE Transactions on biomedical engineering},
  volume={51},
  number={4},
  pages={570--581},
  year={2004},
  publisher={IEEE}
}

@article{makowski2021neurokit2,
  title={NeuroKit2: A Python toolbox for neurophysiological signal processing},
  author={Makowski, Dominique and Pham, Tam and Lau, Zen J and Brammer, Jan C and Lespinasse, Fran{\c{c}}ois and Pham, Hung and Sch{\"o}lzel, Christopher and Chen, SH Annabel},
  journal={Behavior research methods},
  volume={53},
  number={4},
  pages={1689--1696},
  year={2021},
  publisher={Springer}
}

@article{pilia2021ecgdeli,
title = {ECGdeli- An open source ECG delineation toolbox for MATLAB},
journal = {SoftwareX},
volume = {13},
pages = {100639},
year = {2021},
issn = {2352-7110},
doi = {https://doi.org/10.1016/j.softx.2020.100639},
author = {Pilia, Nicolas and Nagel, Claudia and Lenis, Gustavo and Becker, Silvia and Dössel, Olaf and Loewe, Axel},
}

@inproceedings{Emrich2024,
  title={Physiology-Informed ECG Delineation Based on Peak Prominence},
  author={Emrich, Jonas and Gargano, Andrea and Koka, Taulant and Muma, Michael},
  booktitle={2024 32nd European Signal Processing Conference (EUSIPCO)},
  pages={1402--1406},
  year={2024},
  organization={IEEE}
}

@article{cohen2025validation,
  title={Validation of a novel semi-automated ECG quantification tool, applied to a cardio-oncology: Semi-automated ECG Tool applied to cardio-oncology},
  author={Cohen, Samuel D and Robert-Halabi, Maxime and Procureur, Adrien and Jamelot, Mathieu and Vaglio, Martino and Badilini, Fabio and Prifti, Edi and Salem, Joe-Elie},
  journal={Cardio-Oncology},
  volume={11},
  number={1},
  pages={113},
  year={2025},
  publisher={Springer}
}

@article{porretta2025software,
  title={Software-based analysis of T-wave morphology: identifying the electrocardiogram signature of high-risk long QT syndrome},
  author={Porretta, Alessandra Pia and Morgat, Charles and Surget, Elodie and Fressart, V{\'e}ronique and Bloch, Adrien and Neyroud, Nathalie and Badilini, Fabio and Vaglio, Martino and Denjoy, Isabelle and Extramiana, Fabrice},
  journal={Europace},
  volume={27},
  number={9},
  pages={euaf213},
  year={2025},
  publisher={Oxford University Press UK}
}

@article{engstrom2022validation,
  title={Validation of CalECG software for primary prevention heart failure patients: Reducing inter-observer measurement variability},
  author={Engstrom, Nathan and Dobson, Geoffrey P and Ng, Kevin and Lander, Krystle and Win, Kyi and Gupta, Anudeep and Letson, Hayley L},
  journal={Journal of electrocardiology},
  volume={74},
  pages={128--133},
  year={2022},
  publisher={Elsevier}
}

@inproceedings{chen2020simple,
  title={A simple framework for contrastive learning of visual representations},
  author={Chen, Ting and Kornblith, Simon and Norouzi, Mohammad and Hinton, Geoffrey},
  booktitle={International conference on machine learning},
  pages={1597--1607},
  year={2020},
  organization={PmLR}
}

@inproceedings{park2025semisegecg,
  author={Park, Minje and Lim, Jeonghwa and Yu, Taehyung and Joo, Sunghoon},
  title={SemiSegECG: A Multi-Dataset Benchmark for Semi-Supervised Semantic Segmentation in ECG Delineation},
  booktitle={Proceedings of the 34th ACM International Conference on Information and Knowledge Management},
  pages={5099--5104},
  year={2025}
}

@inproceedings{he2022masked,
  title={Masked autoencoders are scalable vision learners},
  author={He, Kaiming and Chen, Xinlei and Xie, Saining and Li, Yanghao and Doll{\'a}r, Piotr and Girshick, Ross},
  booktitle={2022 IEEE/CVF conference on computer vision and pattern recognition (CVPR)},
  pages={15979--15988},
  year={2022},
  organization={IEEE}
}

@inproceedings{he2020momentum,
  title={Momentum contrast for unsupervised visual representation learning},
  author={He, Kaiming and Fan, Haoqi and Wu, Yuxin and Xie, Saining and Girshick, Ross},
  booktitle={2020 IEEE/CVF conference on computer vision and pattern recognition (CVPR)},
  pages={9726--9735},
  year={2020},
  organization={IEEE}
}

@inproceedings{chen2021empirical,
  title={An empirical study of training self-supervised vision transformers},
  author={Chen, Xinlei and Xie, Saining and He, Kaiming},
  booktitle={2021 IEEE/CVF international conference on computer vision (ICCV)},
  pages={9620--9629},
  year={2021},
  organization={IEEE}
}

@article{jaeger2024systematic,
  title={Systematic Comparison of ECG Delineation Algorithm Performance on Smartwatch Data},
  author={Jaeger, Katharina M and Nissen, Michael and Flaucher, Madeleine and Graf, Luisa and Joanidopoulos, Joana and Anneken, Lars and Huebner, Hanna and Goossens, Chlo{\"e} and Titzmann, Adriana and Pontones, Constanza and others},
  journal={IEEE Access},
  volume={12},
  pages={160794--160804},
  year={2024},
  publisher={IEEE}
}

@article{chuma2026performance,
  title={Performance assessment of ECG delineators on single-lead wearable ambulatory data},
  author={Chuma, Amsalu Tomas and Youssef, Ahmed Saeed and Asmare, Melkamu Hunegnaw and Wang, Chunzhuo and Kassie, Desalew Mekonnen and Voigt, Jens-Uwe and Vanrumste, Bart},
  journal={medRxiv},
  pages={2026--03},
  year={2026},
  publisher={Cold Spring Harbor Laboratory Press}
}

@inproceedings{laguna1997database,
  title={A database for evaluation of algorithms for measurement of QT and other waveform intervals in the ECG},
  author={Laguna, Pablo and Mark, Roger G and Goldberg, A and Moody, George B},
  booktitle={Computers in cardiology 1997},
  pages={673--676},
  year={1997},
  organization={IEEE}
}

@article{jimenez2024generalising,
  title={Generalising electrocardiogram detection and delineation: training convolutional neural networks with synthetic data augmentation},
  author={Jimenez-Perez, Guillermo and Acosta, Juan and Alcaine, Alejandro and Camara, Oscar},
  journal={Frontiers in Cardiovascular Medicine},
  volume={11},
  pages={1341786},
  year={2024},
  publisher={Frontiers Media SA}
}

@misc{avetisyan2024ispdataset,
  doi = {10.5281/ZENODO.11472366},
  url = {https://zenodo.org/doi/10.5281/zenodo.11472366},
  author = {Avetisyan,  Aram and Khachaturov,  Nikolas and Asatryan,  Ariana and Tigranyan,  Shahane and Markin,  Yury},
  title = {ISP ECG delineation dataset},
  publisher = {Zenodo},
  year = {2024},
  copyright = {Creative Commons Attribution No Derivatives 4.0 International}
}

@article{wagner2020ptb,
  title={PTB-XL, a large publicly available electrocardiography dataset},
  author={Wagner, Patrick and Strodthoff, Nils and Bousseljot, Ralf-Dieter and Kreiseler, Dieter and Lunze, Fatima I and Samek, Wojciech and Schaeffter, Tobias},
  journal={Scientific data},
  volume={7},
  number={1},
  pages={1--15},
  year={2020},
  publisher={Nature Publishing Group}
}

@article{gow2023mimic,
  title={MIMIC-IV-ECG: Diagnostic Electrocardiogram Matched Subset},
  author={Gow, Brian and Pollard, Tom and Nathanson, Larry A and Johnson, Alistair and Moody, Benjamin and Fernandes, Chrystinne and Greenbaum, Nathaniel and Waks, Jonathan W and Eslami, Parastou and Carbonati, Tanner and others},
  journal={Type: dataset},
  volume={6},
  pages={13--14},
  year={2023}
}

@article{kalyakulina2020ludb,
  title={LUDB: a new open-access validation tool for electrocardiogram delineation algorithms},
  author={Kalyakulina, Alena I and Yusipov, Igor I and Moskalenko, Viktor A and Nikolskiy, Alexander V and Kosonogov, Konstantin A and Osipov, Grigory V and Zolotykh, Nikolai Yu and Ivanchenko, Mikhail V},
  journal={IEEE access},
  volume={8},
  pages={186181--186190},
  year={2020},
  publisher={IEEE}
}

@article{zheng202012,
  title={A 12-Lead ECG database to identify origins of idiopathic ventricular arrhythmia containing 334 patients},
  author={Zheng, Jianwei and Fu, Guohua and Anderson, Kyle and Chu, Huimin and Rakovski, Cyril},
  journal={Scientific data},
  volume={7},
  number={1},
  pages={98},
  year={2020},
  publisher={Nature Publishing Group UK London}
}

@article{RDB202012,
  title={A 12-lead electrocardiogram database for arrhythmia research covering more than 10,000 patients},
  author={Zheng, Jianwei and Zhang, Jianming and Danioko, Sidy and Yao, Hai and Guo, Hangyuan and Rakovski, Cyril},
  journal={Scientific data},
  volume={7},
  number={1},
  pages={48},
  year={2020},
  publisher={Nature Publishing Group UK London}
}

@article{liu2025net,
  title={Y-Net-ECG: A Multi-Lead informed and interpretable architecture for ECG segmentation across diverse rhythms},
  author={Liu, Yuhang and Zhang, Peng and Feng, Xiaoli and Hu, Die and Zhou, Ding and Li, Jingting and Huang, Kaibiao and Zhao, Yinuo and Fu, Zuoming and Zheng, Qianqian and others},
  journal={Expert Systems with Applications},
  volume={283},
  pages={127955},
  year={2025},
  publisher={Elsevier}
}

@article{han2026review,
  title={Review of open foundation models and datasets for ECG and PPG waveforms},
  author={Han, Ahram and Tohyama, Takeshi and Yoon, Dukyong and Paik, Kenneth and Gow, Brian and Celi, Leo Anthony and Lee, Hyeonhoon and Lee, Hyung-Chul},
  journal={npj Digital Medicine},
  year={2026},
  publisher={Nature Publishing Group UK London}
}

@article{chen2023_tol40,
  title={Post-processing refined ECG delineation based on 1D-UNet},
  author={Chen, Zhenqin and Wang, Mengying and Zhang, Meiyu and Huang, Wei and Gu, Hanjie and Xu, Jinshan},
  journal={Biomedical Signal Processing and Control},
  volume={79},
  pages={104106},
  year={2023},
  publisher={Elsevier}
}

@article{Kalyakulina_2019,
   title={Finding Morphology Points of Electrocardiographic-Signal Waves Using Wavelet Analysis},
   volume={61},
   ISSN={1573-9120},
   url={http://dx.doi.org/10.1007/s11141-019-09929-2},
   DOI={10.1007/s11141-019-09929-2},
   number={8-9},
   journal={Radiophysics and Quantum Electronics},
   publisher={Springer Science and Business Media LLC},
   author={Kalyakulina, A. I. and Yusipov, I. I. and Moskalenko, V. A. and Nikolskiy, A. V. and Kozlov, A. A. and Zolotykh, N. Yu. and Ivanchenko, M. V.},
   year={2019},
   month=Jan, pages={689–703} }

@techreport{AAMI_EC57_2020,
  author       = {{Association for the Advancement of Medical Instrumentation}},
  title        = {Testing and Reporting Performance Results of Cardiac Rhythm and {ST} Segment Measurement Algorithms},
  number       = {ANSI/AAMI EC57:1998/(R2020)},
  institution  = {Association for the Advancement of Medical Instrumentation},
  address      = {Arlington, VA, USA},
  year         = {2020}
}

@article{tibshirani1993introduction,
  title={An introduction to the bootstrap},
  author={Tibshirani, Robert J and Efron, Bradley},
  journal={Monographs on statistics and applied probability},
  volume={57},
  number={1},
  pages={1--436},
  year={1993}
}

@article{wilcoxon1945individual,
  title={Individual comparisons by ranking methods},
  author={Wilcoxon, Frank},
  journal={Biometrics bulletin},
  volume={1},
  number={6},
  pages={80--83},
  year={1945},
  publisher={JSTOR}
}

@article{holm1979simple,
  title={A simple sequentially rejective multiple test procedure},
  author={Holm, Sture},
  journal={Scandinavian journal of statistics},
  pages={65--70},
  year={1979},
  publisher={JSTOR}
}

@article{tarvainen2017mean,
  title={Mean teachers are better role models: Weight-averaged consistency targets improve semi-supervised deep learning results},
  author={Tarvainen, Antti and Valpola, Harri},
  journal={Advances in neural information processing systems},
  volume={30},
  year={2017}
}

@inproceedings{
dosovitskiy2021an,
title={An Image is Worth 16x16 Words: Transformers for Image Recognition at Scale},
author={Alexey Dosovitskiy and Lucas Beyer and Alexander Kolesnikov and Dirk Weissenborn and Xiaohua Zhai and Thomas Unterthiner and Mostafa Dehghani and Matthias Minderer and Georg Heigold and Sylvain Gelly and Jakob Uszkoreit and Neil Houlsby},
booktitle={International Conference on Learning Representations},
year={2021},
url={https://openreview.net/forum?id=YicbFdNTTy}
}

@inproceedings{touvron2021training,
  title={Training data-efficient image transformers \& distillation through attention},
  author={Touvron, Hugo and Cord, Matthieu and Douze, Matthijs and Massa, Francisco and Sablayrolles, Alexandre and J{\'e}gou, Herv{\'e}},
  booktitle={International conference on machine learning},
  pages={10347--10357},
  year={2021},
  organization={PMLR}
}

@inproceedings{long2015fully,
  title={Fully convolutional networks for semantic segmentation},
  author={Long, Jonathan and Shelhamer, Evan and Darrell, Trevor},
  booktitle={Proceedings of the IEEE conference on computer vision and pattern recognition},
  pages={3431--3440},
  year={2015}
}

@article{srivastava2014dropout,
  title={Dropout: a simple way to prevent neural networks from overfitting},
  author={Srivastava, Nitish and Hinton, Geoffrey and Krizhevsky, Alex and Sutskever, Ilya and Salakhutdinov, Ruslan},
  journal={The journal of machine learning research},
  volume={15},
  number={1},
  pages={1929--1958},
  year={2014},
  publisher={JMLR. org}
}

@inproceedings{loshchilov2018decoupled,
    title={Decoupled Weight Decay Regularization},
    author={Ilya Loshchilov and Frank Hutter},
    booktitle={International Conference on Learning Representations},
    year={2019},
    url={https://openreview.net/forum?id=Bkg6RiCqY7},
}

@inproceedings{loshchilov2017sgdr,
    title={SGDR: Stochastic Gradient Descent with Warm Restarts},
    author={Ilya Loshchilov and Frank Hutter},
    booktitle={International Conference on Learning Representations},
    year={2017},
    url={https://openreview.net/forum?id=Skq89Scxx}
}

\newpage

\renewcommand{\thetable}{S\arabic{table}}
\renewcommand{\thefigure}{S\arabic{figure}}

\setcounter{table}{0}
\setcounter{figure}{0}
\setcounter{section}{7}

\section{Dataset Details}
\label{sec:supple_dataset_detail}
This section provides further details on the ECG datasets introduced in Section~\ref{sec:datasets} of the main paper and summarized in Table~I. We adopt the standardized benchmark configuration of a prior open benchmark study,\footnote{\url{https://github.com/vuno/semi-seg-ecg}} which publicly released both single-lead ECG recordings and the corresponding split index files. Under this configuration, \textit{QTDB} is partitioned subject-wise into training, validation, and test sets at a 6:2:2 ratio, whereas \textit{ISP} follows its own predefined test split, with the remaining recordings assigned to training and validation at a 6:2 ratio. Table~\ref{tab:labeled_split} reports the sample counts of the internal (\textit{QTDB} $+$ \textit{ISP}) dataset. Tables~\ref{tab:ludb_zhejiang_rhythm} and~\ref{tab:RDB_subgroup} detail the rhythm composition of the external test sets.

\begin{table}[H]
\centering
\setlength{\tabcolsep}{6pt}
\caption{The detail number of the Internal (\textit{QTDB} $+$ \textit{ISP}) set sample.}
\label{tab:labeled_split}
\begin{tabular}{@{}l rrr r@{}}
\toprule
Source & Train & Validation & Test & Total \\
\midrule
\textit{QTDB} & 422 & 148 & 148 & 718 \\
\textit{ISP}  & 3,792 & 1,272 & 924 & 5,988 \\
\midrule
Internal (\textit{QTDB}$+$\textit{ISP}) & 4,214 & 1,420 & 1,072 & 6,706 \\
\bottomrule
\end{tabular}
\end{table}

\begin{table}[H]
\centering
\setlength{\tabcolsep}{5pt}
\renewcommand{\arraystretch}{1.2}
\caption{Rhythm composition of the \textit{LUDB} and \textit{Zhejiang} used for external test.}
\label{tab:ludb_zhejiang_rhythm}
\begin{tabular}{@{}l l r@{}}
\toprule
Dataset & Rhythm & \#ECGs \\
\midrule
\multirow{6}{*}{\textit{LUDB}}
 & Sinus rhythm            & 143 \\
 & Sinus tachycardia       & 4 \\
 & Sinus bradycardia       & 25 \\
 & Sinus arrhythmia        & 8 \\
 & Irregular sinus rhythm  & 2 \\
 & Abnormal rhythm (AFib, flutter) & 18 \\
\cmidrule(lr){2-3}
 & Total                   & 200 \\
\midrule
\multirow{2}{*}{\textit{Zhejiang}}
 & PVC (premature ventricular complex) & 329 \\
 & VT~~(ventricular tachycardia)           & 5 \\
\cmidrule(lr){2-3}
 & Total                   & 334 \\
\bottomrule
\end{tabular}
\end{table}

\begin{table}[H]
\centering
\setlength{\tabcolsep}{5pt}
\renewcommand{\arraystretch}{1.15}
\caption{Rhythm composition of the \textit{RDB} used for the rhythm-stratified
subgroup analysis. Recordings are grouped into sinus and arrhythmia categories.}
\label{tab:RDB_subgroup}
\begin{tabular}{@{}l l r@{}}
\toprule
Group & Rhythm & \#ECGs \\
\midrule
\multirow{4}{*}{Sinus}
 & Sinus rhythm          & 400 \\
 & Sinus bradycardia     & 400 \\
 & Sinus tachycardia     & 140 \\
 & Sinus irregularity    & 399 \\
\midrule
\multirow{4}{*}{Arrhythmia}
 & Atrial flutter            & 400 \\
 & Atrial fibrillation     & 400 \\
 & Atrial tachycardia        & 121 \\
 & Supraventricular tachycardia & 139 \\
\midrule
\multicolumn{2}{@{}l}{Total} & 2{,}399 \\
\bottomrule
\end{tabular}
\end{table}

\section{Evaluation Metrics}
\label{supple_evaluation_metrics_code_detail}
The three complementary metrics used to produce the quantitative results are implemented in the \textbf{Supplementary Code}
: (i) the mean Intersection over Union (mIoU); (ii) the mean absolute error of the PR, QRS, and QT intervals, averaged into a single \textit{averaged interval error}; and (iii) the point-wise sensitivity of the P, QRS, and T onsets and offsets, averaged into a single \textit{averaged point-wise sensitivity}, together with the corresponding positive predictive value (PPV) and localization error, evaluated at tolerances of 40 ms and 150 ms. All metrics are derived from the same predicted segmentation results and share a common set of functions, ensuring that the evaluation is reproducible and applied consistently across datasets.

\subsection{Rhythm-stratified Subgroup Analysis}
We quantify the relative performance degradation from sinus to arrhythmia. For a metric $m$, the degradation is defined as
\[
\Delta = \frac{m_{\text{sinus}} - m_{\text{arr}}}{m_{\text{sinus}}} \times 100\%,
\]
where $m_{\text{sinus}}$ and $m_{\text{arr}}$ denote the value of $m$ on the sinus and arrhythmia subgroups, respectively. For mIoU and averaged point-wise sensitivity, for which higher values indicate better performance, a positive $\Delta$ corresponds to degradation under arrhythmia. For averaged interval error, for which lower values indicate better performance, the sign is reversed,
\[
\Delta = \frac{m_{\text{arr}} - m_{\text{sinus}}}{m_{\text{sinus}}} \times 100\%,
\]
so that a positive $\Delta$ likewise indicates degradation under arrhythmia. Under this convention, a negative $\Delta$ denotes improved performance on the arrhythmia subgroup.

\section{Delineataion Tools}
\label{sec:supple_tool_cfg}

\subsection{Open-source tools}
We use the publicly available implementations of \textsf{NeuroKit2}\footnote{\url{https://github.com/neuropsychology/NeuroKit}}, \textsf{ECGdeli}\footnote{\url{https://github.com/KIT-IBT/ECGdeli}} and \textsf{Prominence}\footnote{\url{https://github.com/JonasEmrich/prominence-delineator}}. All open-source tools are implemented following their documented example configurations.

\subsection{CalECG (AMPS LLC, New York, NY)}
ECG waveforms are analyzed using \textsf{CalECG} (v4.2.0). Because \textsf{CalECG} requires a supported input format, the recordings in each dataset are converted to a \textsf{CalECG}-readable XML format (e.g., GE MUSE XML), preserving the ECG waveform and the dataset's sampling rate; the software then verifies the sampling rate and signal duration of each recording (e.g., 10 s at 500 Hz for \textit{LUDB}). Under \emph{Tools $>$ Set Up Protocol}, the number of beats to measure is set larger than that in any recording, and QRS detection is configured per lead ("QRS detected separately on each lead") to minimize missed detections. After verifying signal quality on the full waveform, beats are labeled semi-automatically according to the protocol (the "Toggles on SemiAutoomatic Mode" button), and the per-beat measurements are exported as a text file for downstream analysis. The \textsf{CalECG} results should be read as an unedited algorithmic baseline rather than the tool's intended-use accuracy. \textsf{CalECG} is intended primarily as an annotation environment, in which the built-in algorithm provides an initial delineation that a human inspects and corrects; we report its output without any such correction. They remain informative nonetheless: among the points it extracts, \textsf{CalECG} performs reasonably on the sinus-dominated \textit{LUDB} but markedly worse on the arrhythmic \textit{Zhejiang}, in both point-wise sensitivity (Table~V) and averaged interval error (Table~\ref{tab:calecg_interval}). This indicates that the initial algorithmic delineation would require substantially more manual correction under arrhythmia than in sinus rhythm.

\section{Model Training Configurations}
\label{supple_implementation_detail}
This section details the semi-supervised and self-supervised configurations used in our experiments. For semi-supervised learning, we adopt the Mean Teacher (MT) framework (Table~\ref{tab:semi-config}). For self-supervised pretraining, we evaluate three algorithms, \textbf{MAE}, \textbf{MoCo}, and \textbf{MERL}; their optimization and architecture-specific hyperparameters are given in Tables~\ref{tab:ssl-optim} and~\ref{tab:ssl-model}, respectively, following each original work adapted to the ECG domain.

\begin{table}[th]
\centering
\setlength{\tabcolsep}{5pt}
\caption{Semi-supervised learning configuration.}
\label{tab:semi-config}
\begin{tabular}{ll}
\toprule
Component & Setting \\
\midrule
Framework & Mean Teacher (MT) \\
Teacher EMA decay & 0.99 \\
Consistency loss weight & 1.0 \\
Weak augmentation & Random resized cropping \\
Strong augmentation & RandAugment($N=3, p=0.5$) \\
\quad Candidate transforms & Powerline noise \\
                          & Sine-wave noise \\
                          & Amplitude scaling \\
                          & White noise \\
\bottomrule
\end{tabular}

\vspace{2pt}
{\footnotesize $N$: number of transforms, $p$: probability}
\end{table}

\begin{table}[th]
\centering
\setlength{\tabcolsep}{5pt}
\caption{Optimization hyperparameters for self-supervised pretraining: \textbf{MAE}, \textbf{MoCo}, and \textbf{MERL}.
All algorithms are trained with the AdamW optimizer and cosine learning rate scheduling.}
\label{tab:ssl-optim}
\begin{tabular}{lccc}
\toprule
Hyperparameter & \textbf{MAE} & \textbf{MoCo} & \textbf{MERL} \\
\midrule
Epochs                     & 800    & 300    & 50 \\
Warm-up epochs             & 40     & 40     & 5 \\
Batch size                 & 2048   & 2048   & 2048 \\
Learning rate         & 1.2e-3 & 1.2e-3 & 2.0e-4 \\
Weight decay               & 0.05   & 0.1    & 1e-5 \\
$(\beta_1, \beta_2)$       & (0.9, 0.95) & (0.9, 0.999) & (0.9, 0.999) \\
\bottomrule
\end{tabular}
\end{table}

\begin{table}[th]
\centering
\small
\setlength{\tabcolsep}{5pt}
\caption{Model architecture and algorithm-specific hyperparameters for self-supervised pretraining.}
\label{tab:ssl-model}
\begin{tabular}{lc}
\toprule
Hyperparameter & Value \\
\midrule
\multicolumn{2}{c}{\textbf{MAE}} \\
\midrule
Masking ratio                 & 0.75 \\
Decoder embedding dimension   & 96 \\
Decoder depth                 & 4 \\
Decoder attention heads       & 3 \\
\midrule
\multicolumn{2}{c}{\textbf{MoCo}} \\
\midrule
Projection hidden dimension   & 4096 \\
Projection output dimension   & 256 \\
Predictor hidden dimension    & 4096 \\
Momentum schedule             & cosine \\
Momentum coefficient (start $\to$ end) & 0.99 $\to$ 1.00 \\
Temperature                   & 0.2 \\
\midrule
\multicolumn{2}{c}{\textbf{MERL}} \\
\midrule
Text encoder                  & MedCPT \\
ECG projection hidden dimension & 256 \\
ECG projection output dimension & 256 \\
Uni-modal alignment dropout   & 0.1 \\
Temperature                   & 0.07 \\
\bottomrule
\end{tabular}
\end{table}

\section{Detailed Quantitative Results}
\label{supple_perfomance_detail}

\subsection{Statistical Analysis and Seed Robustness}
For metrics computed from per-ECG results (mIoU, averaged interval error, averaged point-wise sensitivity), we assess significance by a paired bootstrap over ECG records. For the comparison of pretraining objectives and average ranks, we use paired Wilcoxon signed-rank tests, pairing on dataset$\times$fine-tuning-configuration blocks and on metric$\times$dataset conditions, respectively. All p-values are Holm-corrected over the corresponding family of comparisons.

Tables III and IV in the main paper report results of a single seed. Repeating the same procedure independently for three seeds, both fine-tuning configurations remain significantly superior in 94 of 96 combinations for Table III (2 configurations$\times$4 datasets$\times$3 metrics$\times$4 tools) and in 38 of 40 for Table IV (2 configurations$\times$4 metrics$\times$5 tools), and mIoU is significantly superior in every combination and every seed. All exceptions involve \textsf{Prominence}: averaged point-wise sensitivity on \textit{LUDB} for \textsc{DL delineator} (sup) and averaged interval error on \textit{RDB} for \textsc{DL delineator} (semi) in Table III, and QT interval error on \textit{mECGDB} for both configurations in Table IV. The margins are small. For \textit{LUDB} sensitivity, \textsc{DL delineator} (sup) narrows to +0.5 \%p in its weakest seed, the only seed that does not reach significance. For \textit{mECGDB} QT error, \textsc{DL delineator} (semi) drops to -0.1~ms in its single non-significant seed, while \textsc{DL delineator} (sup) stays between +1.9 and +2.6~ms without reaching significance in any seed. Only the averaged interval error on \textit{RDB} reverses direction, with \textsf{Prominence} more accurate by 0.6~ms in one of the three seeds (+1.3 and +0.2 ms in the other two).

\subsection{Computational Cost}
We measure the computation time of the \textsc{DL delineator} and four rule-based methods (\textsf{NeuroKit2} (CWT), \textsf{NeuroKit2} (DWT), \textsf{ECGdeli}, and \textsf{Prominence}) on \textit{LUDB}, running each method on a single CPU (Intel i7-10750H) with one sample at a time. For every method, we record the time taken to produce the P, QRS, and T fiducial points, and report the mean and standard deviation (SD) over the test set.

The \textsc{DL delineator} is the fastest of the five, runs at a mean of 17.1 ms per sample (Table~\ref{tab:latency}), ahead of \textsf{Prominence} (20.0 ms) and several times faster than the two \textsf{NeuroKit2} variants and \textsf{ECGdeli} (56.3, 63.6, and 141.0 ms). Its runtime is also the most stable (SD 1.9 ms versus 6 ms or more for the others), as a learned model performs a fixed computation regardless of the signal, whereas the search in the rule-based methods scales with waveform complexity. \textsf{NeuroKit2} additionally fails to return an output on 7 samples, which are excluded from its statistics.

\begin{table}[th]
\centering
\caption{Computation time analysis on \textit{LUDB}, reported as a mean $\pm$ standard deviation. The best value is bolded.}
\label{tab:latency}
\begin{tabular}{lc}
\toprule
Method & Time (ms) $\downarrow$ \\
\midrule
\textsf{NeuroKit2} (CWT) & 56.3$\pm$11.9 \\
\textsf{NeuroKit2} (DWT) & 63.6$\pm$11.1 \\
\textsf{ECGdeli} & 141.0$\pm$57.5 \\
\textsf{Prominence} & 20.0$\pm$6.0 \\
\textsc{DL delineator} & \textbf{17.1$\pm$1.9} \\
\bottomrule
\end{tabular}
\end{table}

\subsection{Full Benchmarking Results}
We present the full benchmarking results summarized in the main paper. Tables~\ref{tab:interval} and~\ref{tab:loc-strategy-pointwise} report the averaged interval error and averaged point-wise sensitivity for each combination of pretraining (none, \textbf{MAE}, \textbf{MoCo}, \textbf{MERL}) and fine-tuning (supervised, semi-supervised), where \textbf{MAE} pretraining generally performs best. Table~\ref{tab:calecg_interval} compares the \textsc{DL delineator} with \textsf{CalECG} on averaged interval error (Internal, \textit{LUDB}, \textit{Zhejiang}), reported separately because \textsf{CalECG} was configured to output only the fiducial points required for interval computation. Table~\ref{tab:tol_supp_sens_ppv} additionally compares the \textsc{DL delineator} with existing open-source tools by averaged point-wise sensitivity and PPV at the $40$ and $150$~ms tolerances, and Table~\ref{tab:pointwise_error} reports the per-fiducial localization error ($\mu \pm \sigma$) on the \textit{LUDB} and \textit{Zhejiang} external test sets. Table~\ref{tab:subgroup_supp} presents the rhythm-stratified subgroup analysis on \textit{RDB}, showing that the \textsc{DL delineator} maintains the smallest performance degradation from sinus to arrhythmia.

\begin{table*}[h]
\centering
\setlength{\tabcolsep}{6pt}
\caption{Benchmarking results of learning strategies for ECG delineation. Per-dataset entries are averaged interval error (ms), reported as mean $\pm$ std over three seeds, where Internal denotes the \textit{QTDB} $+$ \textit{ISP} test split. The best value in each column is shown in \textbf{bold}.}
\label{tab:interval}
\begin{tabular}{llccccc}
\toprule
Pretraining & Fine-tuning & Internal & \textit{LUDB} & \textit{Zhejiang} & \textit{mECGDB} & \textit{RDB} \\
\midrule
\multirow{2}{*}{--}
 & Supervised      & $16.2 \pm 0.3$ & $21.1 \pm 0.7$ & $33.1 \pm 2.0$ & $26.1 \pm 0.9$ & $23.4 \pm 0.3$ \\
 & Semi-supervised & $12.8 \pm 0.4$ & $20.8 \pm 2.4$ & $21.7 \pm 3.0$ & $16.2 \pm 0.6$ & $29.8 \pm 4.3$ \\
\midrule
\multirow{2}{*}{\textbf{MAE}}
 & Supervised                  & $\mathbf{11.6 \pm 0.2}$ & $19.2 \pm 1.7$ & $16.5 \pm 1.2$ & $\mathbf{15.0 \pm 0.2}$ & $\mathbf{20.9 \pm 1.2}$ \\
 & Semi-supervised & $11.7 \pm 0.3$ & $18.4 \pm 2.0$ & $\mathbf{15.9 \pm 0.6}$ & $15.7 \pm 0.7$ & $22.7 \pm 1.0$ \\
\midrule
\multirow{2}{*}{\textbf{MoCo}}
 & Supervised      & $12.0 \pm 0.3$ & $21.3 \pm 2.5$ & $21.3 \pm 1.0$ & $15.2 \pm 0.8$ & $22.4 \pm 1.2$ \\
 & Semi-supervised & $11.8 \pm 0.3$ & $\mathbf{17.2 \pm 0.8}$ & $16.9 \pm 0.8$ & $15.4 \pm 0.7$ & $22.3 \pm 0.5$ \\
\midrule
\multirow{2}{*}{\textbf{MERL}}
 & Supervised      & $15.3 \pm 0.2$ & $20.3 \pm 0.7$ & $34.6 \pm 2.5$ & $24.1 \pm 0.7$ & $23.3 \pm 0.4$ \\
 & Semi-supervised & $16.4 \pm 3.5$ & $50.3 \pm 1.4$ & $23.9 \pm 1.7$ & $16.0 \pm 0.9$ & $25.3 \pm 2.9$ \\
\bottomrule
\end{tabular}
\end{table*}

\begin{table*}[h]
\centering
\caption{Benchmarking results of learning strategies for ECG delineation. Per-dataset entries are averaged point-wise sensitivity (\%), reported as mean $\pm$ std over three seeds, where Internal denotes the \textit{QTDB} $+$ \textit{ISP} test split. The best value in each column is shown in \textbf{bold}.}
\label{tab:loc-strategy-pointwise}
\begin{tabular}{llcccc}
\toprule
Pretraining & Fine-tuning & Internal & \textit{LUDB} & \textit{Zhejiang} & \textit{RDB} \\
\midrule
\multirow{2}{*}{--} & Supervised & $91.3 \pm 0.2$ & $82.5 \pm 1.2$& $66.5 \pm 2.6$ & $87.1 \pm 0.7$ \\
                    & Semi-supervised & $95.4 \pm 0.2$ & $86.5 \pm 7.0$  & $76.6 \pm 6.2$ & $90.8 \pm 0.6$ \\
\midrule
\multirow{2}{*}{\textbf{MAE}}  & Supervised    & $96.1 \pm 0.2$ & $86.4 \pm 3.0$ & $85.4 \pm 2.2$ & $\mathbf{93.3 \pm 0.6}$ \\
                      & Semi-supervised   & $\mathbf{96.2 \pm 0.3}$ & $89.2 \pm 3.3$ & $\mathbf{85.9 \pm 1.5}$ & $92.6 \pm 0.3$ \\
\midrule
\multirow{2}{*}{\textbf{MoCo}} & Supervised          & $95.7 \pm 0.1$  & $85.1 \pm 3.5$          & $77.5 \pm 4.2$ & $92.4 \pm 1.0$ \\
                      & Semi-supervised     & $96.1 \pm 0.2$  & $\mathbf{90.3 \pm 3.5}$ & $84.5 \pm 0.7$ & $92.9 \pm 1.0$ \\
\midrule
\multirow{2}{*}{\textbf{MERL}} & Supervised          & $92.4 \pm 0.2$  & $85.8 \pm 0.8$          & $65.9 \pm 0.3$  & $88.6 \pm 0.1$ \\
                      & Semi-supervised     & $95.2 \pm 0.3$  & $76.7 \pm 2.9$          & $72.6 \pm 5.1$  & $91.6 \pm 0.7$ \\
\bottomrule
\end{tabular}
\end{table*}

\begin{table*}[h]
\centering
\setlength{\tabcolsep}{6pt}
\caption{Comparison of the DL delineator with CalECG on averaged interval error. The best value is in \textbf{bold} and the second best is \underline{underlined}.}
\label{tab:calecg_interval}
\begin{tabular}{@{}l ccc@{}}
\toprule
& \multicolumn{3}{c}{Averaged interval error (ms) $\downarrow$}\\
\cmidrule(lr){2-4}
Method & Internal & \textit{LUDB} & \textit{Zhejiang} \\
\midrule
\textsf{CalECG}               & 37.9 & 20.1 & 58.2 \\
\midrule
\textsc{DL delineator} (sup)  & \textbf{11.5} & \underline{19.0}      & \underline{16.3} \\
\textsc{DL delineator} (semi) & \underline{12.0} & \textbf{16.8} & \textbf{15.3} \\
\bottomrule
\end{tabular}
\end{table*}

\begin{table*}[h]
\centering
\setlength{\tabcolsep}{5pt}
\caption{Comparison of the DL delineator with open-source delineation tools, evaluated by averaged point-wise total sensitivity and positive predictive value (PPV) (\%) at tolerances of 40 and 150 ms. The best value in each column is in \textbf{bold}.}
\label{tab:tol_supp_sens_ppv}
\begin{tabular}{@{}ll cc cc@{}}
\toprule
 & & \multicolumn{2}{c}{Sensitivity (\%)} & \multicolumn{2}{c}{PPV (\%)} \\
\cmidrule(lr){3-4} \cmidrule(lr){5-6}
Dataset & Method & 40\,ms & 150\,ms & 40\,ms & 150\,ms \\
\midrule
\multirow{6}{*}{Internal}
 & \textsf{NeuroKit2} (CWT) & 59.4 & 73.7 & 66.1 & 82.1 \\
 & \textsf{NeuroKit2} (DWT) & 60.3 & 88.4 & 60.5 & 88.0 \\
 & \textsf{ECGdeli}         & 82.4 & 97.0 & 77.3 & 90.8 \\
 & \textsf{Prominence}      & 78.1 & 95.9 & 74.2 & 91.1 \\
\cmidrule(lr){2-6}
 & \textsc{DL delineator} (sup)  & \textbf{96.1} & 99.0 & \textbf{95.6} & \textbf{98.6} \\
 & \textsc{DL delineator} (semi) & \textbf{96.1} & \textbf{99.1} & 95.4 & 98.3 \\
\midrule
\multirow{6}{*}{\textit{LUDB}}
 & \textsf{NeuroKit2} (CWT) & 59.9 & 68.2 & 68.5 & 77.9 \\
 & \textsf{NeuroKit2} (DWT) & 62.2 & 88.2 & 62.1 & 87.2 \\
 & \textsf{ECGdeli}         & 80.6 & 97.9 & 71.8 & 86.7 \\
 & \textsf{Prominence}      & 82.9 & 97.3 & 77.7 & 90.9 \\
\cmidrule(lr){2-6}
 & \textsc{DL delineator} (sup)  & 86.6 & 94.5 & 90.4 & 98.8 \\
 & \textsc{DL delineator} (semi) & \textbf{91.1} & \textbf{98.7} & \textbf{91.4} & \textbf{99.0} \\
\midrule
\multirow{6}{*}{\textit{Zhejiang}}
 & \textsf{NeuroKit2} (CWT) & 15.7 & 21.9 & 52.2 & 74.0 \\
 & \textsf{NeuroKit2} (DWT) & 17.8 & 25.2 & 54.3 & 79.6 \\
 & \textsf{ECGdeli}         & 56.3 & 83.5 & 48.2 & 71.1 \\
 & \textsf{Prominence}      & 54.6 & 74.4 & 57.9 & 81.9 \\
\cmidrule(lr){2-6}
 & \textsc{DL delineator} (sup)  & 82.9 & 92.2 & 88.1 & \textbf{97.7} \\
 & \textsc{DL delineator} (semi) & \textbf{86.0} & \textbf{94.9} & \textbf{88.3} & 97.4 \\
\midrule
\multirow{6}{*}{\textit{RDB}}
 & \textsf{NeuroKit2} (CWT) & 54.9 & 66.1 & 60.8 & 73.6 \\
 & \textsf{NeuroKit2} (DWT) & 58.5 & 81.3 & 54.8 & 77.2 \\
 & \textsf{ECGdeli}         & 76.4 & 95.3 & 63.0 & 78.4 \\
 & \textsf{Prominence}      & 73.7 & 87.8 & 66.9 & 79.9 \\
\cmidrule(lr){2-6}
 & \textsc{DL delineator} (sup)  & \textbf{92.6} & 97.5 & \textbf{88.9} & \textbf{93.7} \\
 & \textsc{DL delineator} (semi) & 92.5 & \textbf{98.3} & 87.7 & 93.3 \\
\bottomrule
\end{tabular}
\end{table*}

\begin{table*}[t]
\centering
\footnotesize
\setlength{\tabcolsep}{5pt}
\caption{Comparison of the DL delineator on external test sets (\textit{LUDB} and \textit{Zhejiang}) with open-source and commercial delineation tools, evaluated by point-wise positive predictive value (PPV, \%) per fiducial point. "Total" is the average over the available onset/offset points. The best value is in \textbf{bold}.}
\label{tab:pointwise_ppv}
\begin{tabular}{@{}ll ccccccc@{}}
\toprule
Dataset & Method & P$_{\mathrm{on}}$ & P$_{\mathrm{off}}$ & QRS$_{\mathrm{on}}$ & QRS$_{\mathrm{off}}$ & T$_{\mathrm{on}}$ & T$_{\mathrm{off}}$ & Total \\
\midrule
\multirow{6}{*}{\textit{LUDB}}
 & \textsf{NeuroKit2} (CWT) & 43.5 & 42.8 & 87.4 & 93.8 & 67.5 & 76.0 & 68.5 \\
 & \textsf{NeuroKit2} (DWT) & 62.4 & 53.6 & 75.1 & 92.3 & 32.7 & 56.5 & 62.1 \\
 & \textsf{ECGdeli}         & 55.1 & 66.9 & 96.8 & 84.6 & 63.2 & 64.1 & 71.8 \\
 & \textsf{CalECG}          & 83.8 & - & 95.0 & 96.9 & - & 84.5 & 90.1 \\
 & \textsf{Prominence}      & 64.6 & 65.9 & 91.0 & 84.9 & 79.0 & 80.7 & 77.7 \\
 \cmidrule(lr){2-9}
 & \textsc{DL delineator} (sup)   & 88.1 & \textbf{89.1} & \textbf{99.2} & \textbf{99.0} & \textbf{81.4} & 85.4 & 90.4 \\
 & \textsc{DL delineator} (semi)  & \textbf{92.8} & \textbf{89.1} & \textbf{99.2} & \textbf{99.0} & 81.3 & \textbf{86.8} & \textbf{91.4} \\
\midrule
\multirow{6}{*}{\textit{Zhejiang}}
 & \textsf{NeuroKit2} (CWT) & 34.1 & 30.5 & 77.2 & 83.3 & 41.9 & 46.3 & 52.2 \\
 & \textsf{NeuroKit2} (DWT) & 44.2 & 43.7 & 53.8 & 85.8 & 48.1 & 50.2 & 54.3 \\
 & \textsf{ECGdeli}         & 33.5 & 21.7 & 79.3 & 64.2 & 45.7 & 44.6 & 48.2 \\
 & \textsf{CalECG}          & 30.5 & - & 52.8 & 89.1 & - & 74.0 & 61.6 \\
 & \textsf{Prominence}      & 47.3 & 46.8 & 77.2 & 75.4 & 52.1 & 48.7 & 57.9 \\
 \cmidrule(lr){2-9}
 & \textsc{DL delineator} (sup) & \textbf{92.6} & \textbf{89.6} & 95.8 & 97.1 & \textbf{70.7} & 82.7 & 88.1 \\
 & \textsc{DL delineator} (semi) & 91.1 & 87.6 & \textbf{97.3} & \textbf{97.3} & 70.2 & \textbf{86.2} & \textbf{88.3} \\
\bottomrule
\end{tabular}
\end{table*}

\begin{table*}[h]
\centering
\setlength{\tabcolsep}{6pt}
\caption{Comparison of the DL delineator on \textit{LUDB} and \textit{Zhejiang} external test sets with widely used open-source and commercial delineation tools, evaluated by point-wise localization error (ms) per fiducial point at the 40 ms tolerance. The signed mean error and its standard deviation are reported.}
\label{tab:pointwise_error}
\begin{tabular}{@{}ll cccccc@{}}
\toprule
Dataset & Method & P$_{\mathrm{on}}$ & P$_{\mathrm{off}}$ & QRS$_{\mathrm{on}}$ & QRS$_{\mathrm{off}}$ & T$_{\mathrm{on}}$ & T$_{\mathrm{off}}$ \\
\midrule
\multirow{6}{*}{\textit{LUDB}}
 & \textsf{NeuroKit2} (CWT) & -7.6$\pm$9.4 & 0.1$\pm$9.5 & -8.8$\pm$12.3 & -12.0$\pm$13.3 & 5.0$\pm$17.4 & 6.1$\pm$10.9 \\
 & \textsf{NeuroKit2} (DWT) & 3.5$\pm$17.6 & -3.0$\pm$17.3 & -3.6$\pm$13.7 & -3.1$\pm$13.4 & 17.7$\pm$20.1 & -10.0$\pm$13.2 \\
 & \textsf{ECGdeli} & -13.0$\pm$11.3 & 16.8$\pm$11.4 & -7.2$\pm$13.9 & 16.2$\pm$16.0 & -10.4$\pm$19.9 & 4.5$\pm$15.9 \\
 & \textsf{CalECG} & 0.3$\pm$11.5 & - & 8.1$\pm$13.5 & -2.3$\pm$11.8 & - & 5.3$\pm$12.1 \\
 & \textsf{Prominence} & 1.9$\pm$12.8 & 3.1$\pm$13.9 & -0.0$\pm$15.9 & -11.4$\pm$18.1 & 2.8$\pm$18.2 & 13.2$\pm$14.8 \\
 \cmidrule(lr){2-8}
 & \textsc{DL delineator} (sup) & -5.4$\pm$14.3 & 7.6$\pm$15.1 & -8.4$\pm$10.1 & 10.4$\pm$11.4 & 4.0$\pm$20.1 & 12.5$\pm$14.1 \\
 & \textsc{DL delineator} (semi) & -6.5$\pm$13.1 & 10.0$\pm$14.6 & -8.2$\pm$10.0 & 9.3$\pm$12.1 & 1.3$\pm$20.4 & 13.0$\pm$14.5 \\
\midrule
\multirow{6}{*}{\textit{Zhejiang}}
 & \textsf{NeuroKit2} (CWT) & 6.3$\pm$14.1 & -13.4$\pm$12.8 & -6.8$\pm$15.4 & -12.5$\pm$19.7 & 2.9$\pm$19.3 & -4.6$\pm$16.2 \\
 & \textsf{NeuroKit2} (DWT) & 2.7$\pm$14.4 & -2.9$\pm$16.1 & -5.0$\pm$16.4 & -8.2$\pm$14.5 & 7.6$\pm$20.5 & -7.1$\pm$14.5 \\
 & \textsf{ECGdeli} & -3.9$\pm$12.0 & 6.4$\pm$17.1 & -1.5$\pm$18.2 & 13.3$\pm$18.0 & 1.4$\pm$23.3 & -4.7$\pm$19.5 \\
 & \textsf{CalECG} & 2.8$\pm$16.1 & - & 25.7$\pm$10.7 & -7.2$\pm$13.6 & - & -2.4$\pm$15.1 \\
 & \textsf{Prominence} & 4.8$\pm$13.8 & -4.5$\pm$15.4 & 2.1$\pm$18.9 & -10.3$\pm$17.4 & 3.1$\pm$20.3 & -0.3$\pm$19.3 \\
  \cmidrule(lr){2-8}
 & \textsc{DL delineator} (sup) & -1.1$\pm$11.2 & -5.3$\pm$15.7 & -4.6$\pm$12.9 & 5.5$\pm$13.8 & 8.2$\pm$20.1 & 1.4$\pm$17.5 \\
 & \textsc{DL delineator} (semi) & -0.7$\pm$10.6 & -7.1$\pm$15.3 & -4.5$\pm$13.1 & 4.1$\pm$13.8 & 8.9$\pm$19.9 & -2.3$\pm$17.6 \\
\bottomrule
\end{tabular}%
\end{table*}

\begin{table*}[h]
\centering
\setlength{\tabcolsep}{6pt}
\caption{Rhythm-stratified subgroup analysis on the \textit{RDB}, comparing
the DL delineator with the two strongest tools (Prominence, ECGdeli). Rhythms are grouped into sinus (SR, SB, ST, SI) and arrhythmia (AF, AFIB, AT, SVT). For each metric, we report the group mean performance and its relative degradation (Deg.\%) from sinus to arrhythmia. The best value is in \textbf{bold}.}
\label{tab:subgroup_supp}
\begin{tabular}{@{}ll cccc|c cccc|c c@{}}
\toprule
& & \multicolumn{5}{c}{Sinus group} & \multicolumn{5}{c}{Arrhythmia group} & \\
\cmidrule(lr){3-7}\cmidrule(lr){8-12}
Metric & Method & SR & SB & ST & SI & Avg. & AF & AFIB & AT & SVT & Avg. & Deg.\% \\
\midrule
\multirow{4}{*}{mIoU (\%) $\uparrow$}
 & \textsf{ECGdeli}    & 67.8 & 67.2 & 63.2 & 66.9 & 66.3 & 38.8 & 42.1 & 45.3 & 28.8 & 38.7 & 41.5 \\
 & \textsf{Prominence} & 67.2 & 65.4 & 58.9 & 66.3 & 64.4& 33.1 & 38.1 & 38.6 & 22.8 & 33.2 & 48.6 \\
 & \textsc{DL delineator} (sup)  & 76.4& 77.0& \textbf{75.0}& 75.9& 76.1& \textbf{61.1}& 67.6& \textbf{63.5}& \textbf{67.4}& \textbf{64.9}& \textbf{14.7}\\
 & \textsc{DL delineator} (semi)  & \textbf{77.3}& \textbf{78.0}& 74.2& \textbf{76.9}& \textbf{76.6}& 59.4& \textbf{67.8}& 63.4& 66.3& 64.2& 16.2\\
\midrule
\multirow{4}{*}{Averaged interval error (ms) $\downarrow$}
 & \textsf{ECGdeli}    & 28.3 & 30.1 & 28.4 & 28.6 & 28.9 & \textbf{34.1}& 29.8& 34.4& 62.0& 40.1& \textbf{65.8} \\
 & \textsf{Prominence} & \textbf{17.4} & \textbf{20.1} & \textbf{18.0} & \textbf{17.3} & \textbf{18.2} & 35.7& \textbf{25.7}& 34.9& 56.4& 38.2& 135.6 \\
 & \textsc{DL delineator} (sup)        & 22.0& 21.7& 18.1& 20.5& 20.6& 35.0& 28.4& \textbf{22.9}& \textbf{27.7}& \textbf{28.5}& 71.7\\
 & \textsc{DL delineator} (semi)        & 20.3& \textbf{20.1}& 18.7& 18.4& 19.4& 38.2& 28.5& 24.6& 29.5& 30.2& 92.6\\
\midrule
\multirow{4}{*}{Average point-wise sensitivity (\%) $\uparrow$}
 & \textsf{ECGdeli}    & 82.3 & 79.0 & 80.0 & 83.2 & 81.1 & 70.2 & 74.1 & 69.1 & 55.7 & 67.3 & 17.1 \\
 & \textsf{Prominence} & 86.1 & 82.4 & 79.9 & 85.7 & 83.5 & 59.7 & 69.1 & 57.9 & 46.3 & 58.2 & 30.3 \\
 & \textsc{DL delineator} (sup)       & 97.2& 95.7& \textbf{95.7}& 96.9& 96.4& \textbf{84.4}& \textbf{86.4}& 86.5& 63.0& 80.1& 16.9\\
 & \textsc{DL delineator} (semi)       & \textbf{97.5} & \textbf{96.1}& 95.0& \textbf{97.4}& \textbf{96.5}& 81.8& 86.3& \textbf{88.3}& \textbf{65.1}& \textbf{80.4} & \textbf{16.7}\\
\bottomrule
\end{tabular}%

\vspace{2pt}
{\footnotesize SR: sinus rhythm; SB: sinus bradycardia; ST: sinus tachycardia; SI: sinus irregularity;}

{\footnotesize AF: atrial flutter; AFIB: atrial fibrillation; AT: atrial tachycardia; SVT: supraventricular tachycardia}
\end{table*}

\newpage
\onecolumn 
\section{Qualitative Results}
\label{supple_quality_detail}

\begin{figure*}[h]
    \centering
    \includegraphics[width=1.0\linewidth]{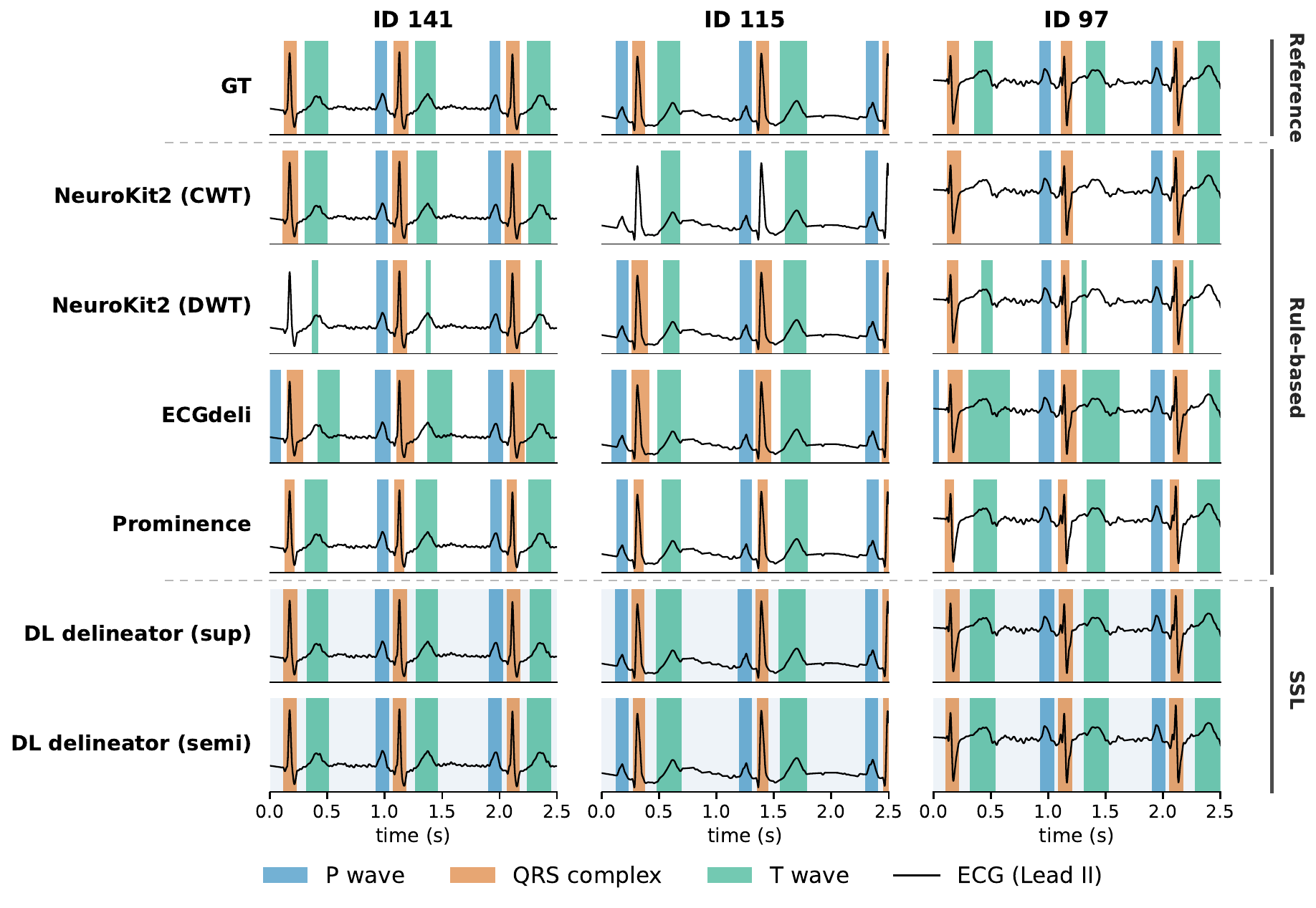}
    \caption{Qualitative P/QRS/T delineation on a representative 2.5\,s Lead II segment from \textit{LUDB}. Rows are grouped into the reference annotation (GT), rule-based tools, and the \textsc{DL delineator}; shaded bands mark each method's detected P (blue), QRS (orange), and T (green) waves on the \emph{same} input signal.}
    \label{fig:quality_ludb_3samples}
\end{figure*}

\begin{figure*}[h]
    \centering
    \includegraphics[width=1.0\linewidth]{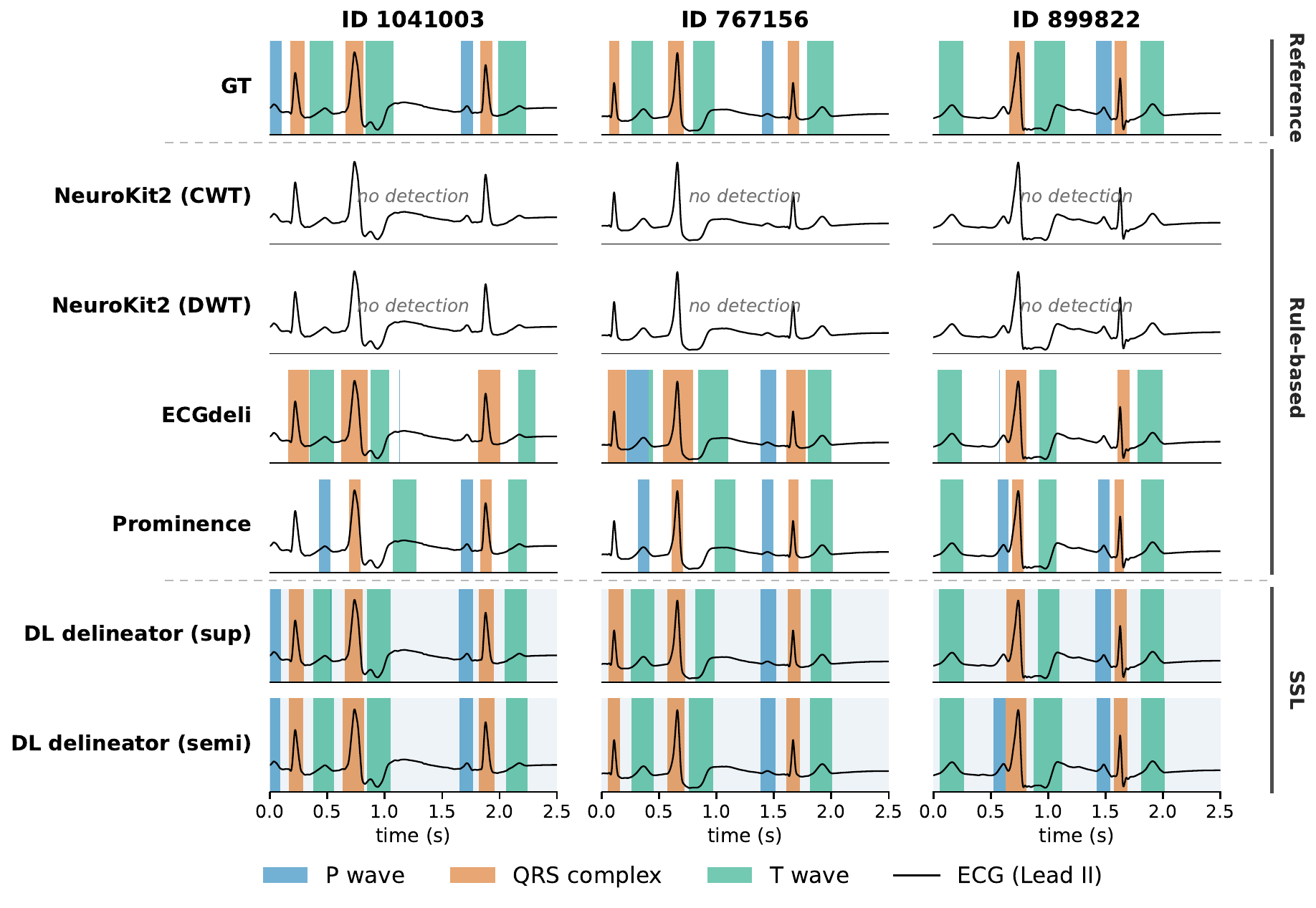}
    \caption{Qualitative P/QRS/T delineation on a representative 2.5\,s Lead II segment from \textit{Zhejiang}. Rows are grouped into the reference annotation (GT), rule-based tools, and the \textsc{DL delineator}; shaded bands mark each method's detected P (blue), QRS (orange), and T (green) waves on the \emph{same} input signal.}
    \label{fig:quality_zhejiang_3samples}
\end{figure*}

\begin{figure*}[h]
    \centering
    \includegraphics[width=1.0\linewidth]{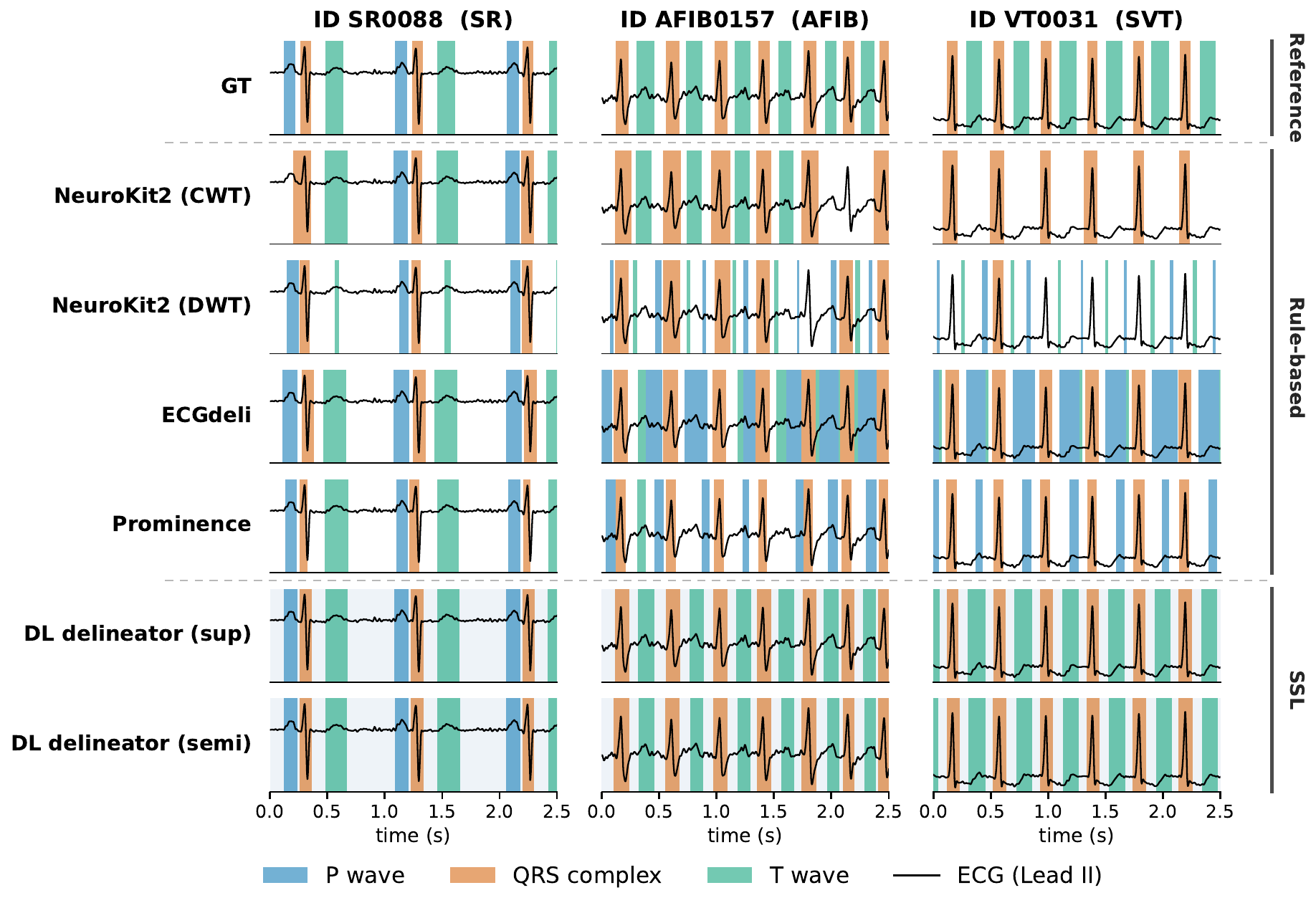}
    \caption{Qualitative P/QRS/T delineation on a representative 2.5\,s Lead II segment from \textit{RDB}. Rows are grouped into the reference annotation (GT), rule-based tools, and the \textsc{DL delineator}; shaded bands mark each method's detected P (blue), QRS (orange), and T (green) waves on the \emph{same} input signal. SR, sinus rhythm; AFIB, atrial fibrillation; SVT, supraventricular tachycardia.}
    \label{fig:quality_rdb_3samples}
\end{figure*}

\end{document}